\documentclass[10pt,a4paper,oneside,notitlepage,fleqn,reqno,english,table]{article}

\usepackage[hidelinks]{hyperref} 
\usepackage{cmap} 				 

\usepackage{multicol} 

\usepackage{longtable} 
\usepackage{multirow} 
\usepackage{rotating} 
\usepackage[cp1252]{inputenc} 
\usepackage{lscape} 
\usepackage{url}    
\usepackage[section]{placeins}  
\usepackage{afterpage} 
\usepackage{capt-of}        
\usepackage{dirtytalk} 

\usepackage[T1]{fontenc}
\usepackage{tabularx, booktabs}   
\usepackage{makecell} 
\usepackage{array} 
\usepackage{xcolor} 
\usepackage{xfrac} 

\usepackage{selinput}
\SelectInputMappings{
  aacute={á},
  eacute={é},
  iacute={í},
  oacute={ó},
  uacute={ú},
  ntilde={ñ}
}

\graphicspath{{figs/}} 					
\usepackage[parfill]{parskip} 				

\usepackage{amsmath} 			
\usepackage{amsfonts}			
\usepackage{mathtools}			
\everymath{\displaystyle}		
\usepackage{amssymb}  			
\usepackage{nicefrac}  			
\usepackage{accents}			

\usepackage{tikz}  
\usetikzlibrary{shapes,arrows,calc,shadows,fit,intersections,datavisualization.formats.functions}

\tikzstyle{block}        = [draw, fill=blue!30, rectangle, text centered, minimum height=3em, minimum width=6em] 
\tikzstyle{blockred}     = [draw, fill=red!30, rectangle, text centered, minimum height=3em, minimum width=6em] 
\tikzstyle{blockyellow}  = [draw, fill=yellow!30, rectangle, text centered, minimum height=3em, minimum width=6em] 
\tikzstyle{blockgreen}   = [draw, fill=green!30, rectangle, text centered, minimum height=3em, minimum width=6em] 
\tikzstyle{blockgrey1}   = [draw, fill=black!8,  rectangle, text centered, minimum height=3em, minimum width=6em] 
\tikzstyle{blockgrey2}   = [draw, fill=black!16, rectangle, text centered, minimum height=3em, minimum width=6em] 
\tikzstyle{blockgrey3}   = [draw, fill=black!24, rectangle, text centered, minimum height=3em, minimum width=6em] 
\tikzstyle{blockgrey4}   = [draw, fill=black!32, rectangle, text centered, minimum height=3em, minimum width=6em] 
\tikzstyle{blockgrey5}   = [draw, fill=black!40, rectangle, text centered, minimum height=3em, minimum width=6em] 
\tikzstyle{blocknofill}  = [draw=black!50, line width=1.5pt, rectangle, rounded corners, text centered, minimum height=3em, minimum width=6em]
\tikzstyle{blockhigh}    = [draw, fill=blue!20, rectangle, text centered, minimum height=6em, minimum width=6em]
\tikzstyle{noblock}      = [draw=black!50, line width=1.5pt, rectangle, rounded corners, text centered, minimum height=3em, minimum width=6em]
\tikzstyle{sum}          = [draw, fill=blue!20, circle, node distance=1cm]
\tikzstyle{sumrest}      = [draw, fill=black, circle, radius=0.5cm]
\tikzstyle{pinstyle}     = [pin edge={to-,thin,black}]
\tikzstyle{title}        = [text centered]
\pgfdeclarelayer{background}
\pgfdeclarelayer{foreground}
\pgfsetlayers{background,main,foreground}

\usepackage{pgfplots}
\pgfplotsset{compat=1.13}
\pgfplotsset{colormap/bluered}
\usepackage{pgfplotstable}
\usepgfplotslibrary{polar}
\usepgfplotslibrary{colormaps}
\usepgfplotslibrary{external}

\usepackage{graphicx, color} 	
\usepackage{epstopdf}  			  

\allowdisplaybreaks 

\newcommand{\ds}    {\displaystyle} 		
\newcommand{\sss}   {\scriptscriptstyle}    

\newcommand{\CC}		{{C\nolinebreak[4]\hspace{-.05em}\raisebox{.4ex}{\tiny\bf ++}}}

\newcommand{\nm}   		[1] {\ensuremath{\mathrm{#1}}} 									
\newcommand{\neweq}     [2] {\begin{equation} \mathrm{#1}\label{#2} \end{equation}} 	

\renewcommand{\vec}		[1] {\mbox{\boldmath{\ensuremath{\mathrm{#1}}}}}	
\newcommand{\derpar}    [2] {\dfrac{d{#1}}{d{#2}}}      					
\newcommand{\lrp}       [1] {\left(#1\right)}								
\newcommand{\lrsb}      [1] {\left[#1\right]}								

\newcommand{\mun}        [1] {\mu_{{#1}n}}
\newcommand{\sigman}     [1] {\sigma_{{#1}n}}
\newcommand{\mui}        [1] {\mu_{{#1}i}}
\newcommand{\sigmai}     [1] {\sigma_{{#1}i}}

\newcommand{\circled}            [1] {\accentset{\circ}{#1}}

\newcommand{\Deltat}     	{\Delta t}
\newcommand{\DeltatTRUTH}	{\Delta t_{\sss TRUTH}}
\newcommand{\DeltatSENSED}  {\Delta t_{\sss SENSED}}
\newcommand{\DeltatEST}  	{\Delta t_{\sss EST}}
\newcommand{\DeltatCNTR}  	{\Delta t_{\sss CNTR}}
\newcommand{\DeltatGNSS}  	{\Delta t_{\sss GNSS}}

\newcommand{\DeltatIMG}  	{\Delta t_{\sss IMG}}

\newcommand{\deltaCNTR}  	{\vec{\delta}_{\sss CNTR}}		
\newcommand{\deltaTARGET}  	{\vec{\delta}_{\sss TARGET}}	    

\newcommand{\xvec}   			{\vec x}

\newcommand{\xvecdot}			{\vec {\dot x}}
\newcommand{\xvecest}	  		{\hat{\vec x}}
\newcommand{\xvecestzero}		{\hat{\vec x}_0}
\newcommand{\xvectilde} 		{\widetilde{\vec x}}

\newcommand{\xvecvis}			{\circled{\vec x}}
\newcommand{\xvecviszero}		{\circled{\vec x}_0}
\newcommand{\lambdatilde}		{\widetilde{\lambda}}
\newcommand{\varphitilde}		{\widetilde{\varphi}}
\newcommand{\htilde}			{\widetilde{h}}

\newcommand{\xTRUTH}			{\vec x_{\sss TRUTH}}
\newcommand{\xSENSED}  			{\vec x_{\sss SENSED}}
\newcommand{\xEST}  			{\vec x_{\sss EST}}
\newcommand{\xIMG}  			{\vec x_{\sss IMG}}
\newcommand{\xREF}	  			{\vec x_{\sss REF}}

\newcommand{\FE}		 {F_{\sss E}}

\newcommand{\iEi}        {\vec i_{1}^{\sss E}}
\newcommand{\iEii}       {\vec i_{2}^{\sss E}}
\newcommand{\iEiii}      {\vec i_{3}^{\sss E}}

\newcommand{\TEgdt}      {\vec T^{\sss E,GDT}}
\newcommand{\TEgdttilde} {\widetilde{\vec T}^{\sss E,GDT}}
\newcommand{\TEgdtdot}   {\vec{\dot T}^{\sss E,GDT}}

\newcommand{\FN} 		  {F_{\sss N}}
\newcommand{\ON}          {O_{\sss N}}                            

\newcommand{\iNi}         {\vec i_{1}^{\sss N}}
\newcommand{\iNii}        {\vec i_{2}^{\sss N}}
\newcommand{\iNiii}       {\vec i_{3}^{\sss N}}

\newcommand{\FB}		  {F_{\sss B}}
\newcommand{\OB}          {O_{\sss B}}                            

\newcommand{\iBi}         {\vec i_{1}^{\sss B}}
\newcommand{\iBii}        {\vec i_{2}^{\sss B}}
\newcommand{\iBiii}       {\vec i_{3}^{\sss B}}

\newcommand{\FI}		 {F_{\sss I}}							

\newcommand{\gcNMODEL}    {\vec g_{c,\sss{MOD}}^{\sss N}}

\newcommand{\vvec} 				{\vec v}

\newcommand{\vB} 	   			{\vec v^{\sss B}}

\newcommand{\vN}	    		{\vec v^{\sss N}} 
\newcommand{\vNi}        		{v_{1}^{\sss N}}
\newcommand{\vNii}      	  	{v_{2}^{\sss N}}
\newcommand{\vNiii} 	      	{v_{3}^{\sss N}}

\newcommand{\vtilde}			{\widetilde{\vec v}}
\newcommand{\vNtilde}			{\widetilde{\vec v}^{\sss N}}
\newcommand{\vNdot}	    		{\vec{\dot v}^{\sss N}}

\newcommand{\vNskew}			{\widehat{\vec v}^{\sss N}}

\newcommand{\vTAS}          {\vec v_{\sss TAS}}    

\newcommand{\vtasINI}       {v_{\sss TAS,INI}}

\newcommand{\vtasEND}       {v_{\sss TAS,END}}

\newcommand{\vwindINI}     	{v_{\sss WIND,INI}}
\newcommand{\vwindEND}     	{v_{\sss WIND,END}}

\newcommand{\omegaE}		{\omega_{\sss E}}

\newcommand{\wIE}			{\vec \omega_{\sss IE}}

\newcommand{\wIEN}			{\vec \omega_{\sss IE}^{\sss N}}

\newcommand{\wIENskew}		{\widehat{\vec \omega}_{\sss IE}^{\sss N}}

\newcommand{\wEN}			{\vec \omega_{\sss EN}}

\newcommand{\wENN}			{\vec \omega_{\sss EN}^{\sss N}}
\newcommand{\wENNskew}		{\widehat{\vec \omega}_{\sss EN}^{\sss N}}

\newcommand{\wNBB}			{\vec \omega_{\sss NB}^{\sss B}}
\newcommand{\wNBBdot}		{\vec{\dot \omega}_{\sss NB}^{\sss B}}

\newcommand{\wIBBtilde}			{\widetilde{\vec \omega}_{\sss IB}^{\sss B}}

\newcommand{\acor}			{\vec a_{cor}}
\newcommand{\acorN}			{\vec a_{cor}^{\sss N}}

\newcommand{\qNB}			{\vec q_{\sss NB}}

\newcommand{\qNBest}		{\hat{\vec q}_{\sss NB}}

\newcommand{\qNBvis}		{\circled{\vec q}_{\sss NB}}

\newcommand{\DeltarNBB}        	{\Delta \vec r_{\sss NB}^{\sss B}}
\newcommand{\DeltarNBBdot}      {\Delta \vec {\dot r}_{\sss NB}^{\sss B}}

\newcommand{\DeltarNBBestnorm}  {\|\Delta \hat{\vec r}_{\sss NB}^{\sss B}\|}

\newcommand{\DeltarNBBvisnorm}  {\|\Delta \circled{\vec r}_{\sss NB}^{\sss B}\|}

\newcommand{\chiWINDINI}       	{\chi_{\sss WIND,INI}}
\newcommand{\chiWINDEND}       	{\chi_{\sss WIND,END}}

\newcommand{\chiINI}            {\chi_{\sss INI}}
\newcommand{\chiEND}            {\chi_{\sss END}}

\newcommand{\xiTURN}            {\xi_{\sss TURN}}

\newcommand{\gammaTASCLIMB}     {\gamma_{\sss TAS,CLIMB}}

\newcommand{\tGNSS}				{t_{\sss GNSS}}
\newcommand{\tEND}				{t_{\sss END}}

\newcommand{\HpINI}         {H_{\sss P,INI}}

\newcommand{\HpEND}         {H_{\sss P,END}}

\newcommand{\hest}			{\hat{h}}
\newcommand{\hvis}			{\circled{h}}

\newcommand{\Deltapest}		{\Delta\hat{p}}
\newcommand{\DeltaTINI}     {\Delta T_{\sss INI}}
\newcommand{\DeltaTEND}     {\Delta T_{\sss END}}

\newcommand{\DeltapINI}     {\Delta p_{\sss INI}}
\newcommand{\DeltapEND}     {\Delta p_{\sss END}}

\newcommand{\fIBB}				{\vec f_{\sss IB}^{\sss B}}

\newcommand{\fIBBtilde}			{\widetilde{\vec f}_{\sss IB}^{\sss B}}

\newcommand{\fIBBdot}			{\vec{\dot f}_{\sss IB}^{\sss B}}

\newcommand{\EACC}		    {\vec E_{\sss ACC}}
\newcommand{\EACCdot}	    {\vec{\dot E}_{\sss ACC}}

\newcommand{\EGYR}			{\vec E_{\sss GYR}}

\newcommand{\EGYRdot}		{\vec{\dot E}_{\sss GYR}}

\newcommand{\BNMODEL}    	{\vec B_{\sss MOD}^{\sss N}}

\newcommand{\BNREAL}    	{\vec B_{\sss REAL}^{\sss N}}

\newcommand{\BNDEV}			{\vec B_{\sss DEV}^{\sss N}}

\newcommand{\BNDEVdot}		{\vec{\dot B}_{\sss DEV}^{\sss N}}

\newcommand{\BBtilde}		{\widetilde{\vec B}^{\sss B}}

\newcommand{\EMAG}	    		{\vec E_{\sss MAG}}
\newcommand{\EMAGdot}	    	{\vec{\dot E}_{\sss MAG}}

\newcommand{\uvec}						{\vec u}
\newcommand{\utilde}					{\widetilde{\vec u}}
\newcommand{\wvec}						{\vec w}
\newcommand{\wtilde}					{\widetilde{\vec w}}
\newcommand{\Avec}						{\vec A}
\newcommand{\Bvec}						{\vec B}

\newcommand{\zvec}						{\vec z}

\newcommand{\Pvec}						{\vec P}
\newcommand{\pvec}						{\vec p}
\newcommand{\yvec}						{\vec y}
\newcommand{\Hvec}						{\vec H}

\newcommand{\Deltaxhorest}				{\Delta \hat{x}_{\sss HOR}}

\newcommand{\Deltaxhorvis}				{\Delta \circled{x}_{\sss HOR}}

\begin{document}

\title{Long Distance GNSS-Denied Visual Inertial Navigation for Autonomous Fixed Wing Unmanned Air Vehicles: SO(3) Manifold Filter based on Virtual Vision Sensor}
\author{Eduardo Gallo\footnote{Contact: e.gallo@alumnos.upm.es, edugallo@yahoo.com, \url{https://orcid.org/0000-0002-7397-0425}} \footnote{Affiliation: Centro de Automática y Robótica, Universidad Politécnica de Madrid - Consejo Superior de Investigaciones Científicas. Centre for Automation and Robotics, Polytechnic University of Madrid.}  and Antonio Barrientos\footnote{Contact: antonio.barrientos@upm.es, \url{https://orcid.org/0000-0003-1691-3907}} \footnote{Affiliation: Centro de Automática y Robótica, Universidad Politécnica de Madrid - Consejo Superior de Investigaciones Científicas. Centre for Automation and Robotics, Polytechnic University of Madrid.}}  
\date{March 2023}
\maketitle


\section*{Abstract}

This article proposes a visual inertial navigation algorithm intended to diminish the horizontal position drift experienced by autonomous fixed wing \texttt{UAV}s (Unmanned Air Vehicles) in the absence of \texttt{GNSS} (Global Navigation Satellite System) signals. In addition to accelerometers, gyroscopes, and magnetometers, the proposed navigation filter relies on the accurate incremental displacement outputs generated by a \texttt{VO} (Visual Odometry) system, denoted here as a Virtual Vision Sensor or \texttt{VVS}, which relies on images of the Earth surface taken by an onboard camera and is itself assisted by the filter inertial estimations. Although not a full replacement for a \texttt{GNSS} receiver since its position observations are relative instead of absolute, the proposed system enables major reductions in the \texttt{GNSS}-Denied attitude and position estimation errors. In order to minimize the accumulation of errors in the absence of absolute observations, the filter is implemented in the manifold of rigid body rotations or \nm{\mathbb{SO}\lrp{3}}. Stochastic high fidelity simulations of two representative scenarios involving the loss of \texttt{GNSS} signals are employed to evaluate the results. The authors release the C++ implementation of both the visual inertial navigation filter and the high fidelity simulation as open-source software \cite{CODE}.


\textbf{\emph{Keywords}}: GNSS-Denied, visual inertial navigation, autonomous navigation, EKF, autonomy, VIO


\section*{Mathematical Notation}

Any variable with a hat accent \nm{< \hat{\cdot} >} refers to its (inertial) estimated value, with a circular accent \nm{< \circled{\cdot} >} to its (visual) estimated value, with a tilde \nm{< \widetilde{\cdot} >} to its measured value, and with a dot \nm{< \dot{\cdot} >} to its time derivative. In the case of vectors, which are displayed in bold (e.g., \nm{\vec x}), other employed symbols include the wide hat \nm{< \widehat{\cdot} >}, which refers to the skew-symmetric form, and the double vertical bars \nm{< \| \cdot \| >}, which refer to the norm. In the case of scalars, the vertical bars \nm{< | \cdot | >} refer to the absolute value. When employing quaternions, which are also displayed in bold (e.g., \nm{\vec q}), the asterisk superindex \nm{< \cdot^{\ast} >} refers to the conjugate, their concatenation and multiplication are represented by \nm{\circ} and \nm{\otimes} respectively, and \nm{\oplus} and \nm{\ominus} refer to the plus and minus operators.

This article makes use of the space of rigid body rotations, and hence relies on the Lie algebra of the special orthogonal group of \nm{\mathbb{R}^3}, known as \nm{\mathbb{SO}(3)}, in particular what refers to the group actions, concatenations, perturbations, and Jacobians, as well its tangent spaces, the  rotation vector \nm{\vec r} and the angular velocity \nm{\vec \omega}. \cite{LIE, Sola2017, Sola2018} are recommended as references.

Four different reference frames are employed in this article: the \texttt{ECEF} frame \nm{\FE}\footnote{The \texttt{ECEF} frame is centered at the Earth center of mass, with \nm{\iEiii} pointing towards the geodetic North along the Earth rotation axis, \nm{\iEi} contained in both the Equator and zero longitude planes, and \nm{\iEii} orthogonal to \nm{\iEi} and \nm{\iEiii} forming a right handed system.}, the \texttt{NED} frame \nm{\FN}\footnote{The \texttt{NED} frame is centered at the aircraft center of mass, with axes aligned with the geodetic North, East, and Down directions.}, the body frame \nm{\FB}\footnote{The body frame is centered at the aircraft center of mass, with \nm{\iBi} contained in the plane of symmetry of the aircraft pointing forward along a fixed direction, \nm{\iBiii} contained in the plane of symmetry of the aircraft, normal to \nm{\iBi} and pointing downward, and \nm{\iBii} orthogonal to both in such a way that they form a right hand system.}, and the inertial frame \nm{\FI}\footnote{The inertial frame is centered at the Earth center of mass and its axes do not rotate with respect to any stars other than the Sun.}. Superindexes are employed over vectors to specify the reference frame in which they are viewed (e.g., \nm{\vN} refers to ground velocity viewed in \texttt{NED}, while \nm{\vB} is the same vector but viewed in body). Subindexes may be employed to clarify the meaning of the variable or vector, such as in \nm{\vTAS} for air velocity instead of the ground velocity \nm{\vec v}, in which case the subindex is either an acronym or its  meaning is clearly explained when first introduced. Subindexes may also refer to a given component of a vector, e.g. \nm{\vNii} refers to the second component of \nm{\vN}. In addition, where two reference frames appear as subindexes to a vector, it means that the vector goes from the first frame to the second. For example, \nm{\wNBB} refers to the angular velocity from \texttt{NED} to body viewed in body.

\section{Introduction and Outline}\label{sec:Intro}

The main objective of this article is to develop a navigation system capable of diminishing the position drift inherent to the flight in \texttt{GNSS} (Global Navigation Satellite System) Denied conditions of autonomous fixed wing aircraft, so in case GNSS signals become unavailable during flight, they have a higher probability of reaching the vicinity of a distant recovery point, from where they can be landed by remote control.

The proposed method combines two different navigation systems (inertial and visual) in such a way that both simultaneously assist and control each other. The inertial attitude and altitude estimations are employed to bound their visual counterparts, resulting in major improvements in the accuracy of the visual horizontal position estimations; these in turn are fed back to the inertial filter in the form of incremental displacement observations, diminishing its estimation errors and partially replacing the \texttt{GNSS} receiver. The resulting horizontal position estimation errors are just a small fraction of what can be obtained by inertial or visual algorithms operating independently. In contrast with most visual inertial methods found in the literature, which focus on the short term \texttt{GNSS}-Denied navigation of ground vehicles, robots, and multirotors, the proposed algorithms are primarily intended for the long distance \texttt{GNSS}-Denied navigation of autonomous fixed wing aircraft.

Section \ref{sec:Novelty} describes the article novelty and main applications. The proposed architecture when combining the inertial and visual navigation algorithms is discussed in section \ref{sec:architecture}. Section \ref{sec:VVS} introduces the \texttt{VVS} (Virtual Vision Sensor), which is just a different name for an advanced visual odometry pipeline assisted by the inertial outputs. The \texttt{VVS} outputs containing the accurate incremental displacement visual estimations are then supplied to a navigation filter implemented in the tangent space to the \nm{\mathbb{SO}\lrp{3}} aircraft attitude; section \ref{sec:VA-INS} contains a detailed explanation of the filter equations. The operation of the fully integrated closed loop visual inertial navigation system is described in section \ref{sec:VINS}.

Section \ref{sec:Simulation} introduces the stochastic high fidelity simulation employed to evaluate the navigation results by means of Monte Carlo executions of two scenarios representative of the challenges of GNSS-Denied navigation. The results obtained when applying the proposed algorithms to these two GNSS-Denied scenarios are described in section \ref{sec:Results}, comparing them with those achieved by standalone inertial and visual systems. Last, the results are summarized for convenience in section \ref{sec:Summary}, while section \ref{sec:Conclusion} provides a short conclusion.

The article concludes with two appendices that are better placed outside the main flow of the document. Appendix \ref{sec:GNSS-Denied} contains an introduction to \texttt{GNSS}-Denied navigation and its challenges, together with reviews of the state-of-the-art in two of the most promising routes to diminish its negative effects, such as \emph{visual odometry} (\texttt{VO}) and \emph{visual inertial odometry} (\texttt{VIO}). Appendix \ref{sec:jacob} describes the obtainment of various Jacobians required for the proposed navigation filter in section \ref{sec:VA-INS}.


\section{Novelty and Application}\label{sec:Novelty}

The proposed approach combines two different navigation systems (inertial and visual) in such a way that both simultaneously assist and control each other, resulting in a virtuous (positive) feed back loop with major improvements in horizontal position estimation accuracy compared with either system by itself.

When compared to the state-of-the-art discussed in appendix \ref{sec:GNSS-Denied}, the proposed system combines the high accuracy representative of tightly coupled smoothers with the lack of complexity and reduced computing requirements of filters and loosely coupled approaches. The existence of two independent solutions (inertial and visual) hints to a loosely coupled approach, but in fact the proposed solution shares most traits with tightly coupled pipelines. First, no information is discarded because the two solutions are not independent, as each simultaneously feeds and is fed by the other. The visual estimations depend on the inertial attitude and altitude outputs, while the inertial filter uses the visual incremental displacement estimations as observations. Unlike loosely coupled solutions, the visual estimations for attitude and altitude are not allowed to deviate from the inertial ones above a certain threshold, so they do not drift. Additionally, the two estimations are never fused together, as the inertial solution constitutes the final output, while the sole role of the visual outputs is to act as the previously mentioned incremental sensor. The proposed solution contains both a filter within the inertial system as well as a keyframe based sliding window smoother within the visual one, obtaining the best properties of both categories.

The \texttt{VIO} solutions listed in appendix \ref{sec:GNSS-Denied} are quite generic with respect to the platforms on which they are mounted, with most applications focused on ground vehicles, indoor robots, and multirotors, as well as with respect to the employed sensors, which are usually restricted to the Inertial Measurement Unit (\texttt{IMU}) and one or more cameras. This article focuses on an specific case (long distance \texttt{GNSS}-Denied turbulent flight of fixed wing autonomous aircraft), and as such it is more restricted in its application, but can also take advantage of the extra sensors already present onboard these platforms (magnetometers, barometer), is better suited to the particular motion of these types of vehicles (forward motion is orders of magnitude faster than those in the lateral and vertical directions), and acts as a solution to \texttt{GNSS}-Denied environments of a different nature than those experienced by those platforms (it can be assumed that \texttt{GNSS} signals are present at the beginning of the flight, and if they disappear, the reason is likely to be technical error or intentional action, so the vehicle needs to be capable of flying for long periods of time in \texttt{GNSS}-Denied conditions).

The proposed system however does not a constitute a full \texttt{GNSS} replacement as it relies on incremental instead of absolute position observations and hence can only reduce the position drift but not eliminate it. Additionally, it imposes certain restrictions on its usage, such as the need for day light, lack of cloud cover below the aircraft, and impossibility to navigate over large bodies of water.

This article proves that inertial and visual navigation systems of a \texttt{UAV} (Unmanned Air Vehicle) can be combined in such a way that the resulting long term \texttt{GNSS}-Denied horizontal position drift is significantly smaller than what can be obtained by either system individually. In the case that \texttt{GNSS} signals become unavailable in mid flight, \texttt{GNSS}-Denied navigation is required for the platform to complete its mission or return to base without the absolute position and ground velocity observations provided by \texttt{GNSS} receivers. As shown in the following sections, the proposed system can significantly increase the possibilities of the \texttt{UAV} safely reaching the vicinity of the intended recovery location, from where it can be landed by remote control. 


\section{Proposed Visual Inertial Navigation Architecture}\label{sec:architecture}

The navigation system processes the measurements of the various onboard sensors (\nm{\xvectilde = \xSENSED}) and generates the estimated state (\nm{\xvecest = \xEST}) required by the guidance and control systems, which contains the best possible estimations of the true aircraft state \nm{\xvec = \xTRUTH}. Note that in this article the images \nm{\vec I} are considered part of the sensed states \nm{\xvectilde = \xSENSED} as the camera is treated as an additional sensor. Hardware and computing power limitations often require the camera and visual navigation algorithms to operate at a lower frequency than those of the sensors and inertial algorithms; table \ref{tab:frequencies} includes the operating frequencies employed in this article.
\begin{center}
\begin{tabular}{lrrcl}
	\hline
	\multicolumn{1}{c}{Discrete Time} & Frequency & Period & Variables & Systems \\
	\hline
	\nm{t_t = t \cdot \DeltatTRUTH}            & \nm{500 \ Hz} & \nm{0.002 \ s} & \nm{\xvec = \xTRUTH}             & Flight physics \\
	\nm{t_s = s \cdot \DeltatSENSED}           & \nm{100 \ Hz} & \nm{0.01 \ s}  & \nm{\xvectilde = \xSENSED}       & Sensors \\ 
	\nm{t_n = n \cdot \DeltatEST}              & \nm{100 \ Hz} & \nm{0.01 \ s}  & \nm{\xvecest = \xEST}            & Inertial navigation \\ 
	\nm{t_c = c \cdot \DeltatCNTR}             & \nm{ 50 \ Hz} & \nm{0.02 \ s}  & \nm{\deltaTARGET, \, \deltaCNTR} & Guidance \& control \\
	\nm{t_i = i \cdot \DeltatIMG}              & \nm{ 10 \ Hz} & \nm{0.1 \ s}   & \nm{\xvecvis = \xIMG}            & Visual navigation \& camera \\
  \nm{t_g = g \cdot \DeltatGNSS}             & \nm{  1 \ Hz} & \nm{  1 \ s}   &                                  & \texttt{GNSS} receiver \\
	\hline
\end{tabular}
\end{center}
\captionof{table}{Working frequencies of the different systems} \label{tab:frequencies}

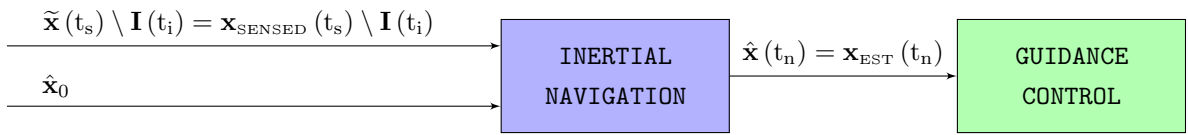
\begin{figure}[h]
\centering
\begin{tikzpicture}[auto, node distance=2cm,>=latex']
	\node [coordinate](midinput) {};
	\node [coordinate, above of=midinput, node distance=0.4cm] (xSENSEDinput){};
	\node [coordinate, below of=midinput, node distance=0.4cm] (xESTzeroinput){};
	\node [block, right of=midinput, minimum width=3.0cm, node distance=8.0cm, align=center, minimum height=1.5cm] (NAVIGATION) {\texttt{INERTIAL} \\ \texttt{NAVIGATION}};
	\node [blockgreen, right of=NAVIGATION, minimum width=3.0cm, node distance=6.0cm, align=center, minimum height=1.5cm] (GC) {\texttt{GUIDANCE} \\ \texttt{CONTROL}};
	\draw [->] (xSENSEDinput) -- node[pos=0.47] {\nm{\xvectilde\lrp{t_s} \setminus \vec I\lrp{t_i} = \xSENSED\lrp{t_s} \setminus \vec I\lrp{t_i}}} ($(NAVIGATION.west)+(0cm,0.4cm)$);
	\draw [->] (xESTzeroinput) -- node[pos=0.1] {\nm{\xvecestzero}} ($(NAVIGATION.west)-(0cm,0.4cm)$);
	\draw [->] (NAVIGATION.east) -- node[pos=0.5] {\nm{\xvecest\lrp{t_n} = \xEST\lrp{t_n}}} (GC.west);
\end{tikzpicture}
\caption{\texttt{INS} flow diagram}
\label{fig:flow_diagram_ins}
\end{figure}

A previous article by the same authors, \cite{INSE}, shows that by freezing the atmospheric and wind estimations at the time the \texttt{GNSS} signals are lost, it is possible to develop an \texttt{EKF} based Inertial Navigation System (\texttt{INS}) for fixed wing aircraft that results in bounded (no drift) estimations for attitude, altitude, and ground velocity\footnote{A bounded attitude estimation ensures that the aircraft can remain aloft in \texttt{GNSS}-Denied conditions for as long as there is fuel available. The altitude estimation error depends on the change in atmospheric pressure offset from its value at the time the \texttt{GNSS} signals are lost, which is bounded by atmospheric physics. The ground velocity estimation error depends on the change in wind velocity from its value at the time the \texttt{GNSS} signals are lost, which is also bounded by atmospheric physics.}, as well as an unavoidable drift in horizontal position caused by integrating the ground velocity without absolute observations. Figure \ref{fig:flow_diagram_ins} graphically depicts that the \texttt{INS} inputs include all sensor measurements with the exception of the camera images \nm{\vec I}.

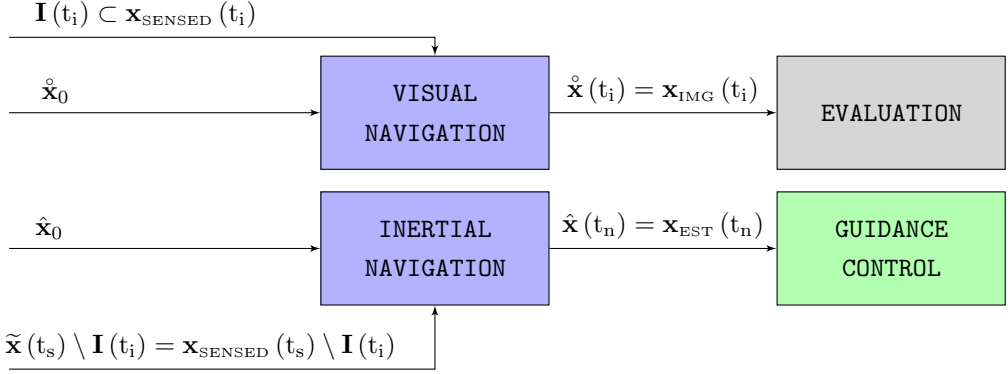
\begin{figure}[h]
\centering
\begin{tikzpicture}[auto, node distance=2cm,>=latex']
	\node [coordinate](midinput) {};
	\node [coordinate, below of=midinput, node distance=0.0cm] (xESTzeroinput0) {};
	\node [coordinate, above of=midinput, node distance=1.8cm] (zetaEBESTzeroinput) {};
	
	\node [coordinate, above of=midinput, node distance=0.4cm] (xSENSEDinput){};
	\node [block, right of=midinput, minimum width=3.0cm, node distance=5.6cm, align=center, minimum height=1.5cm] (NAVIGATION) {\texttt{INERTIAL} \\ \texttt{NAVIGATION}};
	\node [block, above of=NAVIGATION, minimum width=3.0cm, node distance=1.8cm, align=center, minimum height=1.5cm] (VISNAVIGATION) {\texttt{VISUAL} \\ \texttt{NAVIGATION}};
	\node [blockgreen, right of=NAVIGATION, minimum width=3.0cm, node distance=6.0cm, align=center, minimum height=1.5cm] (GC) {\texttt{GUIDANCE} \\ \texttt{CONTROL}};
	\node [blockgrey2, right of=VISNAVIGATION, minimum width=3.0cm, node distance=6.0cm, align=center, minimum height=1.5cm] (EV) {\texttt{EVALUATION}};

	\node [coordinate, right of=NAVIGATION, node distance=4.5cm] (midpoint){};
	\node [coordinate, above of=midpoint, node distance=1.5cm] (pointup){};
	\node [coordinate, below of=midpoint, node distance=1.5cm] (pointdown){};
	\node [coordinate, above of=midpoint, node distance=1.9cm] (xREFinputBIS){};
		
	\draw [->]  (xESTzeroinput0) -- node[pos=0.13] {\nm{\xvecestzero}} (NAVIGATION.west);
	\draw [->] (zetaEBESTzeroinput) -- node[pos=0.15] {\nm{\xvecviszero}} (VISNAVIGATION.west);
	
	\node [coordinate, above of=midpoint, node distance=1.1cm] (uppoint){};
	\draw [->] (NAVIGATION.east) -- node[pos=0.5] {\nm{\xvecest\lrp{t_n} = \xEST\lrp{t_n}}} (GC.west);	
	\draw [->] (VISNAVIGATION.east) -- node[pos=0.5] {\nm{\xvecvis\lrp{t_i} = \xIMG\lrp{t_i}}} (EV.west);
	
	\node [coordinate, below of=midinput, node distance=1.6cm] (xSENSEDINERinput){};
	\node [coordinate, below of=NAVIGATION, node distance=1.6cm] (xSENSEDINERmiddle){};
	\draw [->] (xSENSEDINERinput) -- node[pos=0.45] {\nm{\xvectilde\lrp{t_s} \setminus \vec I\lrp{t_i} = \xSENSED\lrp{t_s} \setminus \vec I\lrp{t_i}}} (xSENSEDINERmiddle) -- (NAVIGATION.south);
	
	\node [coordinate, above of=midinput, node distance=2.8cm] (xSENSEDVISinput){};
	\node [coordinate, above of=VISNAVIGATION, node distance=1.0cm] (xSENSEDVISmiddle){};
	\draw [->] (xSENSEDVISinput) -- (xSENSEDVISmiddle) -- (VISNAVIGATION.north);
	\node at ($(midinput) + (+1.8cm,+3.1cm)$) {\nm{\vec I\lrp{t_i} \subset \xSENSED\lrp{t_i}}};
\end{tikzpicture}
\caption{\texttt{VNS} flow diagram}
\label{fig:flow_diagram_vns}
\end{figure}

A second article, \cite{VNSE}, focuses on visual navigation (that which exclusively relies on the images generated by an onboard camera). Based on the open-source Semi Direct Visual Odometry (\texttt{SVO}) pipeline \cite{Forster2014, Forster2016} introduced in appendix \ref{sec:GNSS-Denied}, it results in a slow but unrestricted error growth or drift in body attitude, altitude, and horizontal position estimations. As excessive attitude errors may create instability issues if they exceed certain thresholds past which the control system is unable to operate as intended, \cite{VNSE} implements the Visual Navigation System (\texttt{VNS}) in an open loop configuration (figure \ref{fig:flow_diagram_vns}), in which the guidance and control systems continue to rely on the inertial (\texttt{INS}) estimations \nm{\xvecest = \xEST}, while the visual states \nm{\xvecvis = \xIMG} are only employed for \texttt{VNS} evaluation.

The strategy to fuse the inertial and visual navigation systems is based on their performances as standalone systems, as summarized in table \ref{tab:compare_standalone}. Inertial systems are clearly superior to visual ones in four out of the six degrees of freedom (attitude and altitude), since their \texttt{GNSS}-Denied estimations are bounded and do not drift. In the case of horizontal position, both systems drift and are hence inappropriate for long term \texttt{GNSS}-Denied navigation, but for different reasons; while the \texttt{INS} drift is the result of integrating the bounded ground velocity estimations without absolute position observations, that of the \texttt{VNS} originates on the slow but continuous accumulation of estimation errors between consecutive images.

It is also worth noting that inertial estimations are significantly noisier than visual ones, so the latter, although worse on an absolute basis, are often more accurate when evaluating the estimations incrementally, this is, from one image to the next or even for the time interval corresponding to a few images.

The fusion between the inertial and visual navigation systems is accomplished in two phases. In the first, the inertial pitch, bank, and altitude bounded estimations are supplied as targets to the \texttt{VNS}, which is now denoted as the Inertially Assisted \texttt{VNS} or \texttt{IA-VNS}, so the visual pitch, bank, and vertical position estimations do not deviate in excess from their inertial counterparts. This process, described in detail in \cite{VNSE}, relies on the addition of priors to the nonlinear pose optimizations within \texttt{SVO}, and does not only result in significantly improved visual estimations for these variables, but more importantly, it also improves the fit between the terrain map built by \texttt{SVO} and the features present in the images, resulting in a major reduction of the horizontal position estimation errors when compared with those achieved by the standalone \texttt{INS} or \texttt{VNS}. Because the drift is reduced but not fully eliminated, the \texttt{IA-VNS} within \cite{VNSE} is also implemented in the open loop configuration depicted in figure \ref{fig:flow_diagram_iavns}, in which the visual outputs are only employed for evaluation purposes.
\begin{center}
\begin{tabular}{lcc}
\hline
\texttt{GNSS}-Denied & Inertial (\texttt{INS}) & Visual (\texttt{VNS}) \\
\hline
Attitude   & Bounded by sensor quality     & Drifts \\
           & Yaw worse than pitch and bank & Yaw better than pitch and bank \\
Vertical   & Bounded by atmospheric physics & Drifts \\
Horizontal & Drifts                        & Drifts \\
\hline
\end{tabular}
\end{center}
\captionof{table}{Summary of inertial and visual standalone navigation systems} \label{tab:compare_standalone}

The \texttt{IA-VNS} qualitative properties are the same as those of the \texttt{VNS} listed in table \ref{tab:compare_standalone}, but quantitatively the drift is significantly lower for five degrees of freedom (yaw accuracy does not change, as discussed in \cite{VNSE}). Of particular interest is the fact that the horizontal position estimations from one image to the next, this is, the incremental displacement observations (equivalent to ground velocity observations), are also even more accurate for the \texttt{IA-VNS} than the already accurate \texttt{VNS} ones, which constitutes the basis of the second phase. 
\begin{figure}[h]
\centering
\begin{tikzpicture}[auto, node distance=2cm,>=latex']
	\node [coordinate](midinput) {};
	\node [coordinate, above of=midinput, node distance=0.0cm] (xESTzeroinput0) {};
	\node [coordinate, above of=midinput, node distance=2.2cm] (xVISSTARzeroinput) {};
	\node [coordinate, above of=midinput, node distance=0.4cm] (xSENSEDinput){};
	
	\node [block, right of=midinput, minimum width=3.0cm, node distance=5.6cm, align=center, minimum height=1.5cm] (NAVIGATION) {\texttt{INERTIAL} \\ \texttt{NAVIGATION}};
	\node [block, above of=NAVIGATION, minimum width=4.5cm, node distance=2.5cm, align=center, minimum height=1.5cm] (VISNAVIGATION) {\texttt{INERTIALLY ASSISTED} \\ \texttt{VISUAL NAVIGATION}};
	\node [blockgreen, right of=NAVIGATION, minimum width=3.0cm, node distance=7.8cm, align=center, minimum height=1.5cm] (GC) {\texttt{GUIDANCE} \\ \texttt{CONTROL}};
	\node [blockgrey2, right of=VISNAVIGATION, minimum width=3.0cm, node distance=7.8cm, align=center, minimum height=1.5cm] (EV) {\texttt{EVALUATION}};
	
	\node [coordinate, right of=NAVIGATION, node distance=2.5cm] (midpoint){};
	\filldraw [black] (midpoint) circle [radius=1pt];
	\node [coordinate, above of=midpoint, node distance=1.2cm] (highpoint){};
		
	\draw [->]  (xESTzeroinput0) -- node[pos=0.08] {\nm{\xvecestzero}} (NAVIGATION.west);
	\draw [->] (xVISSTARzeroinput) -- node[pos=0.10] {\nm{\xvecviszero}} ($(VISNAVIGATION.west)+(0cm,-0.3cm)$);
	
	\draw [->] (NAVIGATION.east) -- node[pos=0.6] {\nm{\xvecest\lrp{t_n} = \xEST\lrp{t_n}}} (GC.west);
	
	\draw [->] (VISNAVIGATION.east) -- node[pos=0.6] {\nm{\xvecvis\lrp{t_i} = \xIMG\lrp{t_i}}} (EV.west);
	
	\draw [->] (midpoint) -- (highpoint) -| (VISNAVIGATION.south);
	
	\node [coordinate, below of=midinput, node distance=1.6cm] (xSENSEDINERinput){};
	\node [coordinate, below of=NAVIGATION, node distance=1.6cm] (xSENSEDINERmiddle){};
	\draw [->] (xSENSEDINERinput) -- node[pos=0.45] {\nm{\xvectilde\lrp{t_s} \setminus \vec I\lrp{t_i} = \xSENSED\lrp{t_s} \setminus \vec I\lrp{t_i}}} (xSENSEDINERmiddle) -- (NAVIGATION.south);
	
	\node [coordinate, above of=midinput, node distance=3.5cm] (xSENSEDVISinput){};
	\node [coordinate, above of=VISNAVIGATION, node distance=1.0cm] (xSENSEDVISmiddle){};
	\draw [->] (xSENSEDVISinput) -- (xSENSEDVISmiddle) -- (VISNAVIGATION.north);
	\node at ($(midinput) + (+1.6cm,+3.1cm)$) {\nm{\vec I\lrp{t_i} \subset \xSENSED\lrp{t_i}}};
	
\end{tikzpicture}
\caption{\texttt{IA-VNS} flow diagram}
\label{fig:flow_diagram_iavns}
\end{figure}

This article proposes an inertial navigation filter implemented on the manifold of rigid body rotations or \nm{\mathbb{SO}\lrp{3}} \cite{LIE, Sola2018}, which in the absence of \texttt{GNSS} signals takes advantage of the \nm{\xvecvis = \xIMG} \texttt{IA-VNS} incremental displacement outputs as if they were the observations generated by an additional sensor denoted as the Virtual Vision Sensor or \texttt{VVS} (section \ref{sec:VVS}). In the proposed \texttt{EKF} (section \ref{sec:VA-INS}), the inertial navigation filter relies on the observations provided by the onboard accelerometers, gyroscopes, and magnetometers, combined with those of the \texttt{GNSS} receiver when signals are available, or those of the \texttt{VVS} when they are not.
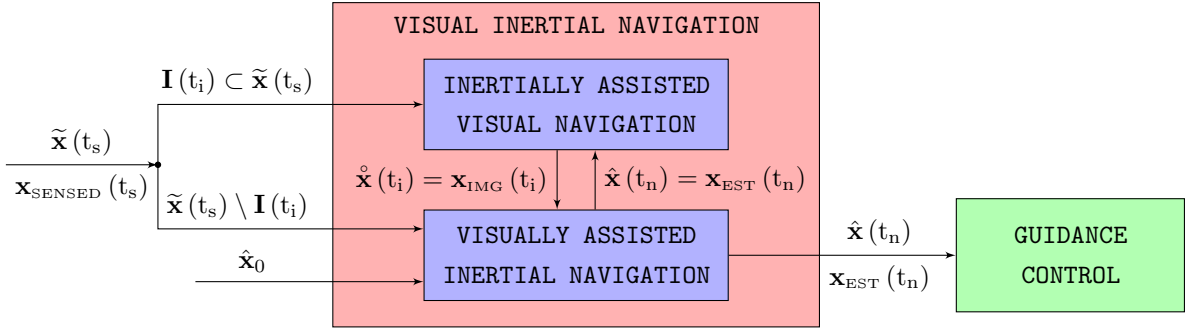
\begin{figure}[h]
\centering
\begin{tikzpicture}[auto, node distance=2cm,>=latex']
	\node [coordinate](midinput) {};
		
	\coordinate (center) at ($(midinput) + (+1.5cm,-0.8cm)$);
	\pgfdeclarelayer{background};
	\node [blockred, right of=center, node distance=5.0cm, minimum width=6.4cm, minimum height=4.3cm] (VINS) {};
	\pgfdeclarelayer{foreground};		
	\node at ($(VINS.north) +(+0.0cm,-0.30cm)$) {\texttt{VISUAL INERTIAL NAVIGATION}};
	
	\node [block, right of=midinput, minimum width=4.0cm, node distance=6.5cm, align=center, minimum height=1.2cm] (VIS NAV) {\texttt{INERTIALLY ASSISTED} \\ \texttt{VISUAL NAVIGATION}};
	\node [block, below of=VIS NAV, minimum width=4.0cm, node distance=2.0cm, align=center, minimum height=1.2cm] (INER NAV) {\texttt{VISUALLY ASSISTED} \\ \texttt{INERTIAL NAVIGATION}};
	\node [blockgreen, right of=INER NAV, minimum width=3.0cm, node distance=6.5cm, align=center, minimum height=1.5cm] (GC) {\texttt{GUIDANCE} \\ \texttt{CONTROL}};
	
	\draw [<-] ($(INER NAV.north)+(-0.25cm,0.0cm)$) -- node[pos=0.5] {\nm{\xvecvis\lrp{t_i} = \xIMG\lrp{t_i}}} ($(VIS NAV.south)+(-0.25cm,0.0cm)$);
	\draw [<-] ($(VIS NAV.south)+(+0.25cm,0.0cm)$) -- node[pos=0.5] {\nm{\xvecest\lrp{t_n} = \xEST\lrp{t_n}}} ($(INER NAV.north)+(+0.25cm,0.0cm)$);
	
	\node [coordinate, right of=INER NAV, node distance=6.7cm] (output){};
	\draw [->] (INER NAV.east) -- (GC.west);
	\node at ($(GC.west) + (-1.0cm,+0.3cm)$) {\nm{\xvecest\lrp{t_n}}};
	\node at ($(GC.west) + (-1.0cm,-0.3cm)$) {\nm{\xEST\lrp{t_n}}};
				
	\coordinate (distrib) at ($(midinput.east) + (+1.0,-0.8cm)$);
	\filldraw [black] (distrib) circle [radius=1pt];	
	\coordinate (xSENSEDinput) at ($(distrib) + (-2.0cm,+0.0cm)$);
	
	\draw [->] (xSENSEDinput) -- (distrib);
	\node at ($(distrib) + (-1.0cm,+0.3cm)$) {\nm{\xvectilde\lrp{t_s}}};
	\node at ($(distrib) + (-1.0cm,-0.3cm)$) {\nm{\xSENSED\lrp{t_s}}};
	
	\draw [->] (distrib) |- node[pos=0.65] {\nm{\vec I\lrp{t_i} \subset \xvectilde\lrp{t_s}}} (VIS NAV.west);
	\draw [->] (distrib) |- node[pos=0.65] {\nm{\xvectilde\lrp{t_s} \setminus \vec I\lrp{t_i}}} ($(INER NAV.west)  + (+0.0cm,+0.35cm)$);
	
	\coordinate (xESTzeroinput) at ($(INER NAV.west)  + (-3.0cm,-0.35cm)$);
	\draw [->] (xESTzeroinput) -- node[pos=0.25] {\nm{\xvecestzero}} ($(INER NAV.west) + (+0.0cm,-0.35cm)$);
\end{tikzpicture}
\caption{\texttt{VINS} flow diagram}
\label{fig:flow_diagram_vins}
\end{figure}

The fully integrated Visual Inertial Navigation System or \texttt{VINS}, graphically depicted in figure \ref{fig:flow_diagram_vins}, discards the original standalone \texttt{INS} and fuses the inertial and visual algorithms so they simultaneously feed and are fed by each other. This creates a positive feed back loop, in which the \texttt{IA-VNS} or \texttt{VVS} horizontal position estimation improvements are successfully transferred to the inertial system (now denoted as the Visually Assisted \texttt{INS} or \texttt{VA-INS}), generating significant reductions in its attitude estimation errors, which in turn generate a second round of major improvements in horizontal position estimation accuracy, as discussed in section \ref{sec:Results}.


\section{Virtual Vision Sensor}\label{sec:VVS}

The \emph{virtual vision sensor} (\texttt{VVS}) constitutes an alternative denomination for the incremental displacement outputs obtained by the \texttt{IA-VNS} \cite{VNSE}. Although the \texttt{IA-VNS} horizontal position estimations are significantly more accurate than those of the standalone \texttt{VNS} (section \ref{sec:Results}), they still drift with time due to the absence of absolute observations. When evaluated on an incremental basis, this is, from one image to the next, the \texttt{IA-VNS} incremental displacement estimations are however quite accurate. This section shows how to convert these incremental estimations into measurements for the geodetic coordinates (\nm{\TEgdttilde}) and the ground velocity viewed in \texttt{NED} (\nm{\vNtilde}), so the \texttt{VVS} can smoothly replace the \texttt{GNSS} receiver within the navigation filter (section \ref{sec:VA-INS}) in the absence of \texttt{GNSS} signals.

Their obtainment relies on the current visual state \nm{\xvecvis\lrp{t_i} = \xIMG\lrp{t_i} = \circled{\vec x}_i} generated by the visual system (\texttt{IA-VNS}), the one obtained with the previous frame \nm{\xvecvis\lrp{t_{i-1}}} = \nm{\xIMG\lrp{t_{i-1}}} = \nm{\circled{\vec x}_{i-1}}, as well as the inertial state \nm{\xvecest\lrp{t_{n - \delta t}}} = \nm{\xEST\lrp{t_{n - \delta t}}} = \nm{\hat{\vec x}_{n - \delta t}} corresponding to the previous image generated by the inertial system (\texttt{VA-INS}). Note that as \nm{t = t_i} = \nm{i \cdot \DeltatIMG} = \nm{t_n = n \cdot \DeltatEST}, the relationship between \emph{i} and \emph{n} is as follows, where \nm{\delta t = 10} is the number of inertial executions for every image being processed (table \ref{tab:frequencies}):
\neweq{n = i \cdot \frac{\DeltatIMG}{\DeltatEST} = i \cdot \delta t}{eq:visual_sensor_delta}

To obtain the \texttt{VVS} velocity (\nm{\vNtilde}) observations, it is necessary to first compute the geodetic coordinates (longitude \nm{\lambda}, latitude \nm{\varphi}, altitude h) derivative as the time difference between their values corresponding to the last two images (\ref{eq:vis_sensor_TEgdtdot}), followed by their transformation into \nm{\vNtilde} per (\ref{eq:vis_sensor_vN})\footnote{M and N represent the \texttt{WGS84} radii of curvature of meridian and prime vertical, respectively.}. Note that \nm{\vNtilde} is very noisy given how it is obtained.
\begin{eqnarray}
\nm{\vec{\dot{\circled T}}_i^{\sss E,GDT}} & = & \nm{\lrsb{\dot{\circled \lambda}_i \ \ \dot{\circled \varphi}_i \ \ \dot{\circled h}_i}^T = \frac{\circled{\vec T}_i^{\sss E,GDT} - \circled{\vec T}_{i-1}^{\sss E,GDT}}{\DeltatIMG}} \label{eq:vis_sensor_TEgdtdot} \\ 
\nm{\widetilde{\vec v}_n^{\sss N}} & = & \nm{\lrsb{\lrsb{M\lrp{{\widetilde \varphi}_{n - \delta t}} + {\widetilde h}_{n - \delta t}} \ \dot{\circled \varphi}_i \ \ \ \ \lrsb{N\lrp{{\widetilde \varphi}_{n - \delta t}} + \widetilde{h}_{n - \delta t}} \ \cos {\widetilde \varphi}_{n - \delta t} \ \dot{\circled \lambda}_i \ \ \ \ - \dot{\circled h}_i}^T} \label{eq:vis_sensor_vN} 
\end{eqnarray}

With respect to the \texttt{VVS} geodetic coordinates, the sensed longitude \nm{\lambdatilde} and latitude \nm{\varphitilde} can be obtained per (\ref{eq:vis_sensor_lambda}) and (\ref{eq:vis_sensor_phi}) based on incrementing the previous inertial estimations (those corresponding to the time of the previous \texttt{VVS} reading) with their visually obtained derivatives. To avoid drift, the geometric altitude \nm{\htilde} is estimated based on the barometer observations assuming that the atmospheric pressure offset remains frozen from the time the \texttt{GNSS} signals are lost\footnote{The interested reader should refer to \cite{INSE} for a detailed explanation on how to estimate the geometric altitude without \texttt{GNSS} signals.}.
\begin{eqnarray}
\nm{\widetilde{\vec T}_n^{\sss E,GDT}} & = & \nm{\lrsb{\widetilde{\lambda}_n \ \ \ \widetilde{\varphi}_n \ \ \ \widetilde{h}_n}^T} \label{eq:vis_sensor_TEdgdt} \\
\nm{\widetilde{\lambda}_n} & = & \nm{\hat{\lambda}_{n - \delta t} + \dot{\circled \lambda}_i \ \DeltatIMG} \label{eq:vis_sensor_lambda} \\
\nm{\widetilde{\varphi}_n} & = & \nm{\hat{\varphi}_{n - \delta t} + \dot{\circled \varphi}_i \ \DeltatIMG} \label{eq:vis_sensor_phi} 
\end{eqnarray}

With respect to the covariances, the authors have assigned the \texttt{VVS} an ad-hoc position standard deviation one order of magnitude lower than that employed for the \texttt{GNSS} receiver so the navigation filter (section \ref{sec:VA-INS}) closely tracks the position observations provided by the \texttt{VVS}\footnote{The \texttt{GNSS} receiver position covariance is higher so the navigation filter slowly corrects the position estimations obtained from integrating the state equations instead of closely adhering to the noisy \texttt{GNSS} position measurements.}. Given the noisy nature of the virtual velocity observations, the authors have preferred to employ a dynamic evaluation for the velocity standard deviation, which coincides with the absolute value (for each of the three dimensions) of the difference between each new velocity observation and the running average of the last twenty readings (equivalent to the last \nm{2 \, s}).	


\section{Proposed Navigation Filter}\label{sec:VA-INS}

As graphically depicted in figure \ref{fig:flow_diagram_vins}, the navigation filter inside the Visually Assisted \texttt{INS} or \texttt{VA-INS} estimates the aircraft pose \nm{\xvecest = \xEST} based on the observations provided by the onboard sensors \nm{\xvectilde \setminus \vec I = \xSENSED \setminus \vec I} complemented if necessary (in the absence of \texttt{GNSS} signals) by the visual observations \nm{\xvecvis = \xIMG} provided by the \texttt{VVS}. Implemented as an \texttt{EKF}, the objective of the navigation filter is the estimation of the aircraft attitude or rotation between the \texttt{NED} and body frames represented by its unit quaternion \nm{\qNB} \cite{LIE, Sola2017} together with the aircraft position represented by its geodetic coordinates \nm{\TEgdt}, although it also estimates the aircraft angular velocity \nm{\wNBB} (from \texttt{NED} to body viewed in body), the ground velocity \nm{\vN} viewed in \texttt{NED}, the specific force \nm{\fIBB} or non gravitational acceleration from the inertial to the body frame viewed in body, the full gyroscope, accelerometer, and magnetometer errors (\nm{\EGYR, \, \EACC, \, \EMAG})\footnote{\nm{\EGYR}, \nm{\EACC}, and \nm{\EMAG} include all gyroscope, accelerometer, and magnetometer error sources except system noise, as explained in \cite{SENSORS}.}, and the difference \nm{\BNDEV} between the Earth magnetic field provided by the onboard model \nm{\BNMODEL} and the real one \nm{\BNREAL}.
\neweq{\xvec\lrp{t} = \lrsb{\qNB \ \ \wNBB \ \ \TEgdt \ \ \vN \ \ \fIBB \ \ \EGYR \ \ \EACC \ \ \EMAG \ \ \BNDEV}^T \hspace{0.2cm}} {eq:nav_vis_iner_filter_st_so3_local_orig}

As the body attitude \nm{\qNB} belongs to the non Euclidean \emph{special orthogonal group} or \nm{\mathbb{SO}(3)}, the classical \texttt{EKF} scheme \cite{LIE, Simon2006}, which relies on linear algebra, shall be avoided in favor of an alternative formulation to ensure that the estimated attitude never deviates from its \nm{\mathbb{SO}(3)} manifold \cite{Blanco2020}\footnote{As \nm{\qNB} is not Euclidean, the direct application of the classical \texttt{EKF} scheme would result in the need to continuously reproject (normalize) the estimated quaternions back to the manifold as otherwise the estimated states would not comply with the constraints (orthogonality and handedness in the case of rigid body rotations). The repeated deviations and reprojections from and to the manifold may result in a significant degradation in the estimation accuracy \cite{LIE, Sola2018}.}. The use of Lie theory, with its manifolds and tangent spaces, enables the construction of rigorous calculus techniques to handle uncertainties, derivatives, and integrals of non Euclidean elements with precision and ease \cite{Sola2018}.

The most rigorous and precise way to adapt the \texttt{EKF} scheme to the presence of non Euclidean elements such as rigid body rotations (\nm{\qNB}) is to exclude the rotation element \nm{\mathcal R \in \mathbb{SO}\lrp{3}} from the state system, replacing the unit quaternion \nm{\qNB} by a local (body) tangent space perturbation represented by the rotation vector \nm{\DeltarNBB \in \mathbb{R}^3} \cite{LIE,Sola2018}. Note that the state vector also contains the velocity of \nm{\mathcal R} as it moves along its manifold (its angular velocity), contained in the local or body tangent space. Each filter step now consists on estimating the rotation element \nm{\hat{\mathcal R}_n = \hat{\mathcal R}\lrp{t_n} \in \mathbb{SO}\lrp{3}}, the state vector \nm{\xvecest_n = \xvecest\lrp{t_n} = \lrsb{\Delta \hat{\vec r}_{{\sss NB}n}^{\sss B} \ \ \hat{\vec \omega}_{{\sss NB}n}^{\sss B} \ \ \hat{\vec z}_n}^T \in \mathbb{R}^{27}}, and its covariance \nm{\Pvec_n = \Pvec\lrp{t_n} \in \mathbb{R}^{27x27}}, based on their values at \nm{t_{n-1} = \lrp{n-1} \, \Deltat}. 
\begin{eqnarray}
\nm{\mathcal R\lrp{t}} & = & \nm{\qNB \hspace{0.2cm} \in \mathbb{SO}(3)\lrp{t}} \label{eq:nav_vis_iner_filter_X_so3_local} \\
\nm{\xvec\lrp{t}}      & = & \nm{\lrsb{\DeltarNBB \ \ \wNBB \ \ \zvec}^T = \lrsb{\DeltarNBB \ \ \pvec}^T} \nonumber \\
                       & = & \nm{\lrsb{\DeltarNBB \ \ \wNBB \ \ \TEgdt \ \ \vN \ \ \fIBB \ \ \EGYR \ \ \EACC \ \ \EMAG \ \ \BNDEV}^T \hspace{0.2cm} \in \mathbb{R}^{27}\lrp{t}}\label{eq:nav_vis_iner_filter_st_so3_local} 
\end{eqnarray}

The observations \nm{\yvec_n} are provided by the gyroscopes, which measure the angular velocity \nm{\wIBBtilde} from the inertial frame to the body frame viewed in body, the accelerometers that provide the specific force \nm{\fIBBtilde} or non gravitational acceleration from the inertial to the body frame viewed in body, the magnetometers that measure the magnetic field \nm{\BBtilde} in the body frame, and either the \texttt{GNSS} receiver or the \texttt{VVS} (section \ref{sec:VVS}) that provide the absolute position (geodetic coordinates) and \texttt{NED} velocity observations (\nm{\TEgdttilde, \, \vNtilde}). Note that the measurements are provided at different frequencies, as listed in table \ref{tab:frequencies}; most are available every \nm{\DeltatSENSED = 0.01 \ s}, but the ground velocity and position measurements are generated every \nm{\DeltatGNSS = 1 \ s} if provided by the \texttt{GNSS} receiver or every \nm{\DeltatIMG = 0.1 \ s} when the \texttt{GNSS} signals are not available and the readings are instead supplied by the \texttt{VVS}.
\begin{eqnarray}
\nm{\yvec_n}           & = & \nm{\lrsb{\wIBBtilde \ \ \fIBBtilde \ \ \BBtilde \ \ \TEgdttilde \ \ \vNtilde}^T \hspace{0.2cm} \in {\mathbb{R}_n^{15}}}\label{eq:nav_vis_iner_filter_y_so3_local}
\end{eqnarray}

The variation with time of the state variables (\ref{eq:nav_vis_iner_cont_time_system_so3_local}) comprises a continuous time nonlinear state system composed by three exact equations (with no simplifications) containing the rotational motion time derivative (\ref{eq:nav_vis_iner_filter_DeltarNBBdot_so3_local}) \cite{LIE, Sola2018}, the kinematics of the geodetic coordinates (\ref{eq:nav_vis_iner_filter_TEgdtdot_so3_local})\footnote{M and N represent the \texttt{WGS84} radii of curvature of meridian and prime vertical, respectively.}, as well as the aircraft velocity dynamics (\ref{eq:nav_vis_iner_filter_vNdot_so3_local}) \cite{INSE}, together with six other differential equations (\ref{eq:nav_vis_iner_filter_other_so3_local}) in which the filter does not posses any knowledge about the time derivatives of the state variables, which are set to zero. Note that in expression (\ref{eq:nav_vis_iner_filter_vNdot_so3_local}), \nm{\vec g_{*}()} represents the \nm{\mathbb{SO}\lrp{3}} group rotation action, \nm{\oplus} constitutes the \nm{\mathbb{SO}\lrp{3}} plus operator, the wide hat \nm{< \widehat{\cdot} >} refers to the skew-symmetric form of a vector, \nm{\wENN} is the motion angular velocity\footnote{The motion angular velocity \nm{\wEN} represents the rotation experienced by any object that moves without modifying its attitude with respect to the Earth surface.} viewed in \texttt{NED}, \nm{\gcNMODEL} constitutes the gravity acceleration modeled by the navigation system, and \nm{\acorN} represents the Coriolis acceleration also viewed in \texttt{NED}. In expression (\ref{eq:nav_vis_iner_cont_time_system_so3_local}) \nm{\vec u} is the known control or input vector and \nm{\vec w} is the process noise\footnote{Note that although present, system noise \nm{\wvec\lrp{t}} is not shown in equations (\ref{eq:nav_vis_iner_filter_DeltarNBBdot_so3_local}) through (\ref{eq:nav_vis_iner_filter_other_so3_local}).}.
\begin{eqnarray}
\nm{\xvecdot\lrp{t}} & = & \nm{\lrsb{\DeltarNBBdot \ \ \wNBBdot \ \ \vec {\dot z}}^T = \vec f\big(\qNB\lrp{t} \oplus \DeltarNBB\lrp{t}, \, \wNBB\lrp{t}, \, \vec z\lrp{t}, \, \uvec\lrp{t}, \ \wvec\lrp{t}, \, t\big)} \label{eq:nav_vis_iner_cont_time_system_so3_local} \\
\nm{\vec {\dot x}_{1-3}}   & = & \nm{\DeltarNBBdot = \wNBB = {\vec x}_{4-6}} \label{eq:nav_vis_iner_filter_DeltarNBBdot_so3_local} \\  
\nm{\vec {\dot x}_{7-9}}   & = & \nm{\TEgdtdot = \lrsb{\dfrac{\vNii}{\lrsb{N\lrp{\varphi} + h} \, \cos\varphi} \ \ \ \ \dfrac{\vNi}{M\lrp{\varphi} + h} \ \ \ \ - \vNiii}^T} \label{eq:nav_vis_iner_filter_TEgdtdot_so3_local} \\
\nm{\vec {\dot x}_{10-12}} & = & \nm{\vNdot = \vec g_{{\ds{\qNB \oplus \DeltarNBB}}*}\lrp{\fIBB} - \wENNskew \; \vN + \gcNMODEL - \acorN} \label{eq:nav_vis_iner_filter_vNdot_so3_local} \\  
\nm{\vec {\dot x}_{OTHER}} & = & \nm{\lrsb{\wNBBdot \ \ \fIBBdot \ \ \EGYRdot \ \ \EACCdot \ \ \EMAGdot \ \ \BNDEVdot}^T = \vec{0}_{18}} \label{eq:nav_vis_iner_filter_other_so3_local} \\
\nm{\uvec\lrp{t}}	       & = & \nm{\vec 0} \label{eq:nav_vis_iner_filter_u_so3_local} 
\end{eqnarray}

Its linearization results in the following system matrix \nm{\vec A\lrp{t} = \partial{\vec f} / \partial{\vec x} \lrp{\vec w = \vec 0} \in \mathbb{R}^{27x27}}, in which appendix \ref{sec:jacob} describes the obtainment of the various Jacobians. Note that the linearization neglects the influence of the geodetic coordinates \nm{\TEgdt} on four different terms (\nm{\TEgdtdot}, \nm{\wENN}, \nm{\gcNMODEL}, and \nm{\acorN}) present in the (\ref{eq:nav_vis_iner_cont_time_system_so3_local}) differential equations, but includes the influence of the aircraft velocity \nm{\vN} on three of them (gravity does not depend on velocity). This is because the \nm{\vN} variation amount that can be experienced in a single state estimation step can have a significant influence on the values of the variables considered, while that of the geodetic coordinates \nm{\TEgdt} does not and hence can be neglected.
\begin{eqnarray}
\nm{\xvecdot\lrp{t}} & = & \nm{\Avec\lrp{t} \, \xvec\lrp{t} + \Bvec\lrp{t} \, \utilde\lrp{t} + \wtilde\lrp{t}} \label{eq:nav_vis_iner_filter_stdot_so3_local} \\
\nm{\vec A_{1-3,4-6}}     & = & \nm{\derpar{\DeltarNBBdot}{\wNBB} = \vec I_{3x3}} \label{eq:nav_vis_iner_filter_A_DeltarNBBdot_wNBB_so3_local} \\
\nm{\vec A_{7-9,10-12}}   & = & \nm{\derpar{\TEgdtdot}{\vN} = \vec J_{\ds{\; \vN}}^{\ds{\; \vec {\dot{T}}^{\sss E,GDT}}}} \label{eq:nav_vis_iner_filter_A_TEgdt_vN_so3_local} \\ 
\nm{\vec A_{10-12,1-3}}   & = & \nm{\derpar{\vNdot}{\DeltarNBB} = \vec J_{\ds{\; \mathcal R_{\sss NB}}}^{\ds{\; \vec g_{\qNB*}(\fIBB)}}} \label{eq:nav_vis_iner_filter_A_vNdot_DeltarNBB_so3_local} \\
\nm{\vec A_{10-12,10-12}} & = & \nm{\derpar{\vNdot}{\vN} = - \wENNskew + \vNskew \, \vec J_{\ds{\; \vN}}^{\ds{\; \wENN}} - \vec J_{\ds{\; \vN}}^{\ds{\; \acorN}}} \label{eq:nav_vis_iner_filter_A_vNdot_vN_so3_local} \\
\nm{\vec A_{10-12,13-15}} & = & \nm{\derpar{\vNdot}{\fIBB} = \vec J_{\ds{\; \fIBB}}^{\ds{\; \vec g_{\qNB \oplus \DeltarNBB*}(\fIBB)}}} \label{eq:nav_vis_iner_filter_A_vNdot_fIBB_so3_local} \\
\nm{A_{ij,OTHER}}         & = & \nm{0} \label{eq:nav_vis_iner_filter_A_other_so3_local} \\
\nm{\utilde\lrp{t}}       & = & \nm{\vec 0} \label{eq:nav_vis_iner_filter_utilde_so3_local}
\end{eqnarray}

The discrete time nonlinear observations system (\ref{eq:nav_vis_iner_filter_yn_so3_local}) contains the observations provided by the different sensors, without any simplifications. New terminology includes the \nm{\mathbb{SO}\lrp{3}} group adjoint \nm{\vec{Ad}()}, inverse adjoint \nm{\vec{Ad}^{-1}()}, and inverse rotation \nm{\vec g_{*}^{-1}()} actions. In addition, \nm{\wIEN} represents the Earth angular velocity (viewed in \texttt{NED}) caused by its rotation around the \nm{\iEiii} axis at a constant rate \nm{\omegaE}, and \nm{\vvec_n} represents the measurement or observation noise\footnote{Note that although present, measurement noise \nm{\vvec_n} is not shown in equations (\ref{eq:nav_vis_iner_filter_wIBBtilde_so3_local}) through (\ref{eq:nav_vis_iner_filter_vNtilde_so3_local}).}.
\begin{eqnarray}
\nm{\yvec_n}     & = & \nm{\vec h\lrp{\vec q_{{\sss NB}n} \oplus \Delta \vec r_{{\sss NB}n}^{\sss B}, \, \vec \omega_{{\sss NB}n}^{\sss B}, \, \vec z_n, \, \, \vvec_n, \, t_n}}\label{eq:nav_vis_iner_filter_yn_so3_local} \\
\nm{\wIBBtilde}  & = & \nm{\wNBB + \vec {Ad}_{{\ds{\vec q_{\sss NB} \oplus \DeltarNBB}}}^{-1} \lrp{\wIEN + \wENN} + \EGYR}\label{eq:nav_vis_iner_filter_wIBBtilde_so3_local} \\
\nm{\fIBBtilde}  & = & \nm{\fIBB + \EACC} \label{eq:nav_vis_iner_filter_fIBBtilde_so3_local} \\
\nm{\BBtilde}    & = & \nm{\vec g_{{\ds{\vec q_{\sss NB} \oplus \DeltarNBB}}*}^{-1} \lrp{\BNMODEL - \BNDEV} + \EMAG} \label{eq:nav_vis_iner_filter_BBtilde_so3_local} \\
\nm{\TEgdttilde} & = & \nm{\TEgdt} \label{eq:nav_vis_iner_filter_TEgdttilde_so3_local} \\
\nm{\vNtilde}    & = & \nm{\vN} \label{eq:nav_vis_iner_filter_vNtilde_so3_local}
\end{eqnarray}

Its linearization results in the following output matrix \nm{\vec H_n = \partial{\vec h} / \partial{\xvec_n}\lrp{\vvec_n = \vec 0} \in \mathbb{R}^{15x27}}, in which appendix \ref{sec:jacob} describes the obtainment of the various Jacobians. As in the state system case above, the linearization also neglects the influence of the geodetic coordinates \nm{\TEgdt} on three different terms (\nm{\wIEN}, \nm{\wENN}, and \nm{\BNMODEL}) present in the (\ref{eq:nav_vis_iner_filter_yn_so3_local}) observation equations. Note that it is not necessary to evaluate the observations input vector \nm{\zvec_n} as it plays no role in the solution \cite{LIE, Simon2006}.
\begin{eqnarray}
\nm{\yvec_n} & \nm{\approx} & \nm{\Hvec_n \, \xvec_n + \zvec_n + \vtilde_n}\label{eq:nav_vis_iner_filter_obser_so3_local} \\
\nm{\vec H_{1-3,1-3}}	  & = & \nm{\derpar{\wIBBtilde}{\DeltarNBB} = \vec J_{\ds{\; \mathcal R_{\sss NB}}}^{\ds{\; \vec {Ad}_{\qNB}^{-1}(\wIEN + \wENN)}}} \label{eq:nav_vis_iner_filter_H_wIBBtilde_DeltarNBB_so3_local} \\
\nm{\vec H_{1-3,4-6}}	  & = & \nm{\derpar{\wIBBtilde}{\wNBB} = \vec{I}_{3x3}} \label{eq:nav_vis_iner_filter_H_wIBBtilde_wNBB_so3_local} \\
\nm{\vec H_{1-3,10-12}}	  & = & \nm{\derpar{\wIBBtilde}{\vN} = \vec J_{\ds{\; \wENN}}^{\ds{\; \vec {Ad}_{\qNB \oplus \DeltarNBB}^{-1}(\wENN)}} \ \vec J_{\ds{\; \vN}}^{\ds{\; \wENN}}} \label{eq:nav_vis_iner_filter_H_wIBBtilde_vN_so3_local} \\
\nm{\vec H_{1-3,16-18}}	  & = & \nm{\derpar{\wIBBtilde}{\EGYR} = \vec{I}_{3x3}} \label{eq:nav_vis_iner_filter_H_wIBBtilde_EGYR_so3_local} \\
\nm{\vec H_{4-6,13-15}}   & = & \nm{\derpar{\fIBBtilde}{\fIBB} = \vec{I}_{3x3}} \label{eq:nav_vis_iner_filter_H_fIBBtilde_fIBB_so3_local} \\
\nm{\vec H_{4-6,19-21}}   & = & \nm{\derpar{\fIBBtilde}{\EACC} = \vec{I}_{3x3}} \label{eq:nav_vis_iner_filter_H_fIBBtilde_EACC_so3_local} \\
\nm{\vec H_{7-9,1-3}}     & = & \nm{\derpar{\BBtilde}{\DeltarNBB} = \vec J_{\ds{\; \mathcal R_{\sss NB}}}^{\ds{\; \vec g_{\qNB*}^{-1}(\BNMODEL - \BNDEV)}}} \label{eq:nav_vis_iner_filter_H_BBtilde_DeltarNBB_so3_local} \\
\nm{\vec H_{7-9,22-24}}   & = & \nm{\derpar{\BBtilde}{\EMAG} = \vec{I}_{3x3}} \label{eq:nav_vis_iner_filter_H_BBtilde_EMAG_so3_local} \\
\nm{\vec H_{7-9,25-27}}   & = & \nm{\derpar{\BBtilde}{\BNDEV} = - \ \vec J_{\ds{\; \BNDEV}}^{\ds{\; \vec g_{\qNB \oplus \DeltarNBB*}^{-1}(\BNDEV)}}} \label{eq:nav_vis_iner_filter_H_BBtilde_BNDEV_so3_local} \\
\nm{\vec H_{10-12,7-9}}   & = & \nm{\derpar{\TEgdttilde}{\TEgdt} = \vec{I}_{3x3}} \label{eq:nav_vis_iner_filter_H_TEgdttilde_TEgdt_so3_local} \\
\nm{\vec H_{13-15,10-12}} & = & \nm{\derpar{\vNtilde}{\vN} = \vec{I}_{3x3}} \label{eq:nav_vis_iner_filter_H_vNtilde_vN_so3_local} \\
\nm{H_{ij,OTHER}}         & = & \nm{0} \label{eq:nav_vis_iner_filter_H_other_so3_local} 
\end{eqnarray}

Based on the above linearized continuous state system and discrete observations, each step within the the classical \texttt{EKF} implementation \cite{LIE, Simon2006} results in estimations for both the state vector \nm{\xvecest_n = \lrsb{\Delta \hat{\vec r}_{{\sss NB}n}^{\sss B} \ \ \hat{\vec \omega}_{{\sss NB}n}^{\sss B} \ \ \hat{\vec z}_n}^T} as well as its covariance \nm{\Pvec_n}. Note that the definition of the covariance matrix is a combination of that of its Euclidean components and its local Lie counterparts \cite{LIE}, with additional combined members\footnote{\nm{\ominus} represents the \nm{\mathbb{SO}\lrp{3}} rotation group minus operator \cite{LIE,Sola2018}.}:
\begin{eqnarray}
\nm{\Pvec_n} & = & \nm{\begin{bmatrix} \nm{\vec C_{{\mathcal {RR}},n}^{\mathcal R}} & \nm{\vec C_{{\mathcal R}p,n}^{\mathcal R}} \\ \nm{\vec C_{p{\mathcal R},n}^{\mathcal R}} & \nm{\vec C_{pp,n}} \end{bmatrix} \ \ \ \ \ \ \ \ \ \ \ \ \ \ \ \ \ \ \ \ \ \ \ \ \ \ \ \ \ \ \ \ \ \ \ \ \ \ \ \ \ \ \ \ \ \ \ \ \ \ \ \ \ \ \ \ \ \ \ \ \ \ \ \ \ \, \in \mathbb{R}^{27x27}} \label{eq:algebra_SS_EKF_P_generic} \\
\nm{\vec C_{{\mathcal {RR}},n}^{\mathcal R}} & = & \nm{E\lrsb{\Delta \vec r_{{\sss NB}n}^{\sss B} \, \Delta \vec r_{{\sss NB}n}^{{{\sss B},T}}} = E\lrsb{\lrp{\mathcal R_{{\sss NB}n} \ominus \vec \mu_{{\mathcal R},{\sss NB}n}} \, \lrp{\mathcal R_{{\sss NB}n} \ominus \vec \mu_{{\mathcal R},{\sss NB}n}}^T} \ \ \ \ \ \ \ \ \ \in \mathbb{R}^{3x3}} \label{eq:algebra_SS_EKF_P_lie} \\
\nm{\vec C_{{\mathcal R}p,n}^{\mathcal R}} & = & \nm{E\lrsb{\Delta \vec r_{{\sss NB}n}^{\sss B} \, \lrp{\vec p_n - \vec \mu_{p,n}}^T} = E\lrsb{\lrp{\mathcal R_{{\sss NB}n} \ominus \vec \mu_{{\mathcal R},{\sss NB}n}} \, \lrp{\vec p_n - \vec \mu_{p,n}}^T} \ \ \ \ \ \ \, \in \mathbb{R}^{3x24}} \label{eq:algebra_SS_EKF_P_lieeuc} \\
\nm{\vec C_{p{\mathcal R},n}^{\mathcal R}} & = & \nm{E\lrsb{\lrp{\vec p_n - \vec \mu_{p,n}} \, \Delta \vec r_{{\sss NB}n}^{{{\sss B}},T}} = E\lrsb{\lrp{\vec p_n - \vec \mu_{p,n}} \, \lrp{\mathcal R_{{\sss NB}n} \ominus \vec \mu_{{\mathcal R},{\sss NB}n}}^T} \ \ \ \ \ \ \ \ \ \in \mathbb{R}^{24x3}} \label{eq:algebra_SS_EKF_P_euclie} \\
\nm{\vec C_{pp,n}} & = & \nm{E\lrsb{\lrp{\vec p_n - \vec \mu_{p,n}} \, \lrp{\vec p_n - \vec \mu_{p,n}}^T} \ \ \ \ \ \ \ \ \ \ \ \ \ \ \ \ \ \ \ \ \ \ \ \ \ \ \ \ \ \ \ \ \ \ \ \ \ \ \ \ \ \ \ \ \ \ \ \ \ \  \in \mathbb{R}^{24x24}} \label{eq:algebra_SS_EKF_P_euc} 
\end{eqnarray}

In the modified \texttt{EKF} scheme suggested here, each estimation step concludes by resetting the estimation of the tangent space perturbation \nm{\Delta \hat{\vec r}_{{\sss NB}n}^{\sss B}} to zero while modifying the estimations for the Lie rotation group element \nm{\hat{\mathcal R}_{{\sss NB}n}} and the error covariance \nm{\Pvec_n} accordingly\footnote{The estimations for angular velocity \nm{\hat{\vec \omega}_{{\sss NB}n}^{\sss B}} and the Euclidean components \nm{\hat{\vec z}_n} are not modified.}. Note that the accuracy of the linearizations required to obtain the \nm{\vec A} and \nm{\vec H} system and output matrices are based on first order Taylor expansions, which are directly related to the size of the tangent space perturbations. Although it is not strictly necessary to reset the perturbations in every \texttt{EKF} cycle, the accuracy of the whole state estimation process is improved by maintaining the perturbations as small as possible, so it is recommended to never bypass the reset step.

Taking into account that the rotation element is going to be updated per (\ref{eq:nav_vis_iner_filter_qNB_reset_so3_local}), the error covariance is propagated to the new rotation element as follows \cite{LIE}, where the Jacobian is provided by (\ref{eq:jac_lie1}) within appendix \ref{sec:jacob}:
\begin{eqnarray}
\nm{\Pvec_n} & \nm{\longleftarrow} & \nm{\vec D \, \Pvec_n \, \vec D^T} \label{eq:algebra_SS_P_reset} \\
\nm{\vec D} & = & \nm{\begin{bmatrix} \nm{\vec J_{\ds{\; \mathcal R}{\sss NB}}^{\ds{\; \hat{\mathcal R}_{{\sss NB}n} \oplus \Delta \hat{\vec r}_{{\sss NB}n}^{\sss B}}}} & \nm{\vec{0}_{3x24}} \\ \nm{\vec{0}_{24x3}} & \nm{\vec {I}_{24x24}} \end{bmatrix} \ \ \ \ \in \mathbb{R}^{27x27}} \label{eq:algebra_SS_D_reset} 
\end{eqnarray}

Last, the rotation element is propagated by means of the local rotation vector perturbation, which is itself reset to zero:
\begin{eqnarray}
\nm{\hat{\vec q}_{{\sss NB}n}} & \nm{\longleftarrow} & \nm{\hat{\vec q}_{{\sss NB}n} \oplus \Delta \hat{\vec r}_{{\sss NB}n}^{\sss B}} \label{eq:nav_vis_iner_filter_qNB_reset_so3_local} \\
\nm{\Delta \hat{\vec r}_{{\sss NB}n}^{\sss B}} & \nm{\longleftarrow} & \nm{\vec{0}_3} \label{eq:nav_vis_iner_filter_DeltarNB_reset_so3_local} 
\end{eqnarray}

Different position (geodetic coordinates) and ground velocity system noise values are employed depending on whether the measurements are provided by the \texttt{GNSS} receiver or the \texttt{VVS}. Lower geodetic coordinates system noise values are employed when \texttt{GNSS} signals are available, as the objective is for the solution to avoid position jumps by smoothly following the state equations and only slightly updating the position based on the \texttt{GNSS} observations to avoid any position drift on the long term. When the position observations are instead supplied by the \texttt{VVS}, higher system noise values are employed so the \texttt{EKF} relies more on the observations and less on the integration. Note that the \texttt{VVS} velocity observations are very noisy because of (\ref{eq:vis_sensor_TEgdtdot}), so it is better if the \texttt{EKF} closely adheres to the position observations that originate at the \texttt{IA-VNS}, correcting in each step as necessary.


\section{Fully Integrated Visual Inertial Navigation System}\label{sec:VINS}

The fully integrated \texttt{VINS} combines the visual (\texttt{IA-VNS}) and inertial (\texttt{VA-INS}) systems in such a way that each assists the other while being simultaneously controlled by it, as depicted in figure \ref{fig:flow_diagram_vins}. As the \texttt{VINS} needs to estimate the state \nm{\xvecest = \xEST} at every \nm{t_n = t_s = n \cdot \DeltatEST = s \cdot \DeltatSENSED}, it has three different operation modes, depending not only on whether \texttt{GNSS} signals are available or not, but also on whether there exists a position and ground velocity observation at the execution time \nm{t_n = t_s}:
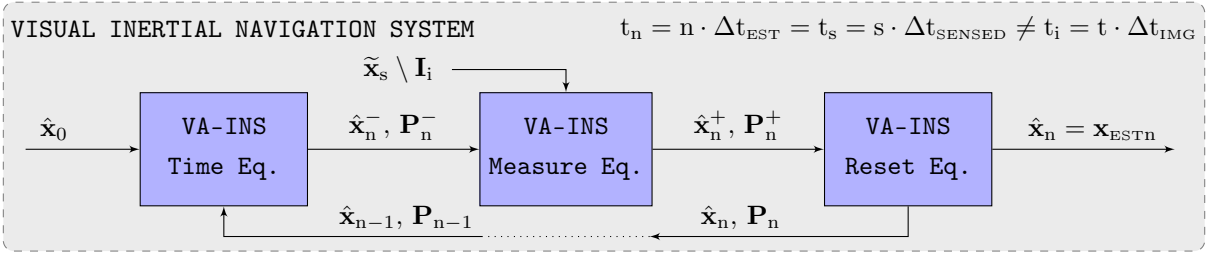
\begin{figure}[h]
\centering
\begin{tikzpicture}[auto, node distance=2cm,>=latex']
	\node [coordinate](midinput) {};
	
	\coordinate (center) at ($(midinput) + (+0.0cm,+0.3cm)$);
	\pgfdeclarelayer{background};
	\node [rectangle, draw=black!50, fill=black!8, dashed, rounded corners, right of=center, node distance=5.0cm, minimum width=15.8cm, minimum height=3.3cm] (VINS) {};
	\pgfdeclarelayer{foreground};

	\node [block, right of=midinput, minimum width=2.2cm, node distance=0.0cm, align=center, minimum height=1.5cm] (TIME)  {\texttt{VA-INS} \\ \texttt{Time Eq.}};
	\node [block, right of=TIME, minimum width=2.2cm, node distance=4.5cm, align=center, minimum height=1.5cm] (MEAS)  {\texttt{VA-INS} \\ \texttt{Measure Eq.}};
	\node [block, right of=MEAS, minimum width=2.2cm, node distance=4.5cm, align=center, minimum height=1.5cm] (RESET) {\texttt{VA-INS} \\ \texttt{Reset Eq.}};
	
	\draw [->] ($(TIME.west)+(-1.5cm,+0.0cm)$) -- node[pos=0.25] {\nm{\xvecestzero}} (TIME.west);
	\draw [->] (TIME.east) -- node[pos=0.5] {\nm{\xvecest^-_n, \, \Pvec^-_n}} (MEAS.west);
	\draw [->] (MEAS.east) -- node[pos=0.5] {\nm{\xvecest^+_n, \, \Pvec^+_n}} (RESET.west);
	\draw [->] (RESET.east) -- node[pos=0.55] {\nm{\xvecest_n = \vec x_{{\sss EST}n}}} ($(RESET.east)+(+2.4cm,+0.0cm)$);
	
	\draw [->] ($(MEAS.north) + (-1.5cm,+0.30cm)$) -| (MEAS.north);
	\node at ($(MEAS.north) +(-2.2cm,+0.30cm)$) {\nm{\xvectilde_s \setminus \vec I_i}};	
		
	\coordinate (ResetOut) at ($(RESET.south) + (-3.4cm,-0.40cm)$);
	\coordinate (TimeIn) at ($(TIME.south)  + (+3.4cm,-0.40cm)$);
	
	\draw [->] (RESET.south) |- (ResetOut);
	\draw [dotted] (ResetOut) -- (TimeIn);
	\draw [->] (TimeIn) -| (TIME.south);
	
	\node at ($(RESET.south) +(-2.2cm,-0.15cm)$) {\nm{\xvecest_n, \, \Pvec_n}};
	\node at ($(TIME.south) +(+2.4cm,-0.15cm)$) {\nm{\xvecest_{n-1}, \, \Pvec_{n-1}}};		
	
	\node at ($(VINS.west) +(+3.15cm,+1.30cm)$) {\texttt{VISUAL INERTIAL NAVIGATION SYSTEM}};
	\node at ($(VINS.east) +(-3.90cm,+1.30cm)$) {\nm{t_n = n \cdot \DeltatEST = t_s = s \cdot \DeltatSENSED \neq t_i = t \cdot \DeltatIMG}};
\end{tikzpicture}
\caption{\texttt{VINS} flow diagram when \nm{t_n = t_s \neq t_i}}
\label{fig:VINS_flow_diagram_times_equal_not}
\end{figure}

\begin{itemize}
\item Given the operating frequencies of the different sensors (table \ref{tab:frequencies}), most \texttt{VINS} executions (ninety-nine out of one hundred for \texttt{GNSS}-Based navigation, nine out of ten in case of \texttt{GNSS}-Denied conditions) can not rely on the position and ground velocity observations provided by either the \texttt{GNSS} receiver or the \texttt{VVS}. As graphically depicted in figure \ref{fig:VINS_flow_diagram_times_equal_not}, these \texttt{VINS} executions rely on the proposed navigation filter (section \ref{sec:VA-INS}), which follows the modified Lie algebra \texttt{EKF} scheme consisting on the successive application (after initialization) of its time update, measurement update, and reset equations\footnote{As discussed in section \ref{sec:VA-INS}, the reset step is not present in a fully Euclidean \texttt{EKF} scheme.}. Note that the filter observation system (\ref{eq:nav_vis_iner_filter_yn_so3_local}) needs to be modified to remove (\ref{eq:nav_vis_iner_filter_TEgdttilde_so3_local}) and (\ref{eq:nav_vis_iner_filter_vNtilde_so3_local}) as there are no \texttt{GNSS} receiver or \texttt{VVS} observations available. 
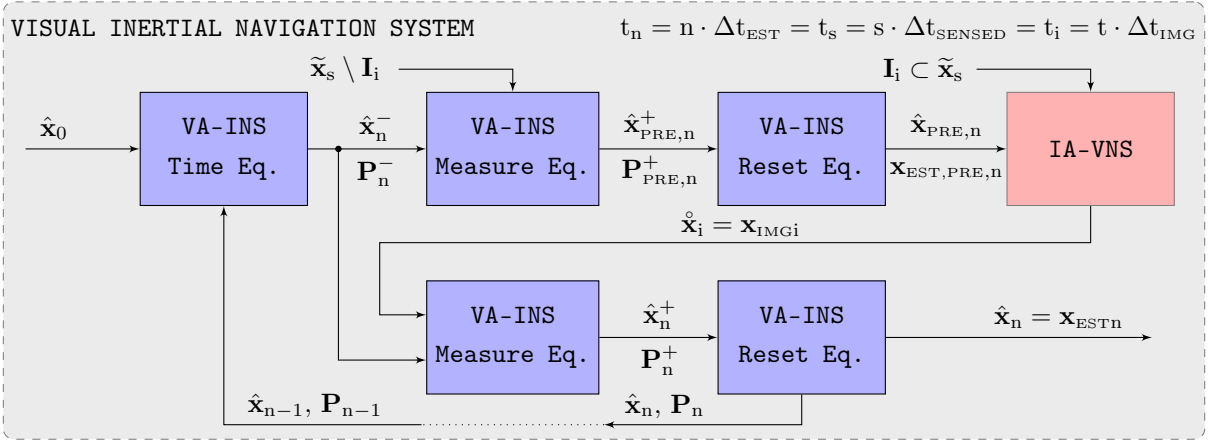
\begin{figure}[h]
\centering
\begin{tikzpicture}[auto, node distance=2cm,>=latex']
	\node [coordinate](midinput) {};
	
	\coordinate (center) at ($(midinput) + (+0.0cm,-0.95cm)$);
	\pgfdeclarelayer{background};
	\node [rectangle, draw=black!50, fill=black!8, dashed, rounded corners, right of=center, node distance=5.0cm, minimum width=15.8cm, minimum height=5.8cm] (VINS) {};
	\pgfdeclarelayer{foreground};		
				
	\node [block, right of=midinput, minimum width=2.2cm, node distance=0.0cm, align=center, minimum height=1.5cm] (TIME)  {\texttt{VA-INS} \\ \texttt{Time Eq.}};
	\node [block, right of=TIME, minimum width=2.2cm, node distance=3.8cm, align=center, minimum height=1.5cm] (MEAS1)  {\texttt{VA-INS} \\ \texttt{Measure Eq.}};
	\node [block, right of=MEAS1, minimum width=2.2cm, node distance=3.8cm, align=center, minimum height=1.5cm] (RESET1) {\texttt{VA-INS} \\ \texttt{Reset Eq.}};
	\node [rectangle, draw=black!50, fill=red!30, right of=RESET1, minimum width=2.2cm, node distance=3.8cm, align=center, minimum height=1.5cm] (VIS)  {\texttt{IA-VNS}};
	
	\draw [->] ($(TIME.west)+(-1.5cm,+0.0cm)$) -- node[pos=0.25] {\nm{\xvecestzero}} (TIME.west);
	\draw [->] (TIME.east) -- (MEAS1.west);
	\node at ($(TIME.east) +(+0.9cm,+0.3cm)$) {\nm{\xvecest^-_n}};
	\node at ($(TIME.east) +(+0.9cm,-0.3cm)$) {\nm{\Pvec^-_n}};
	\draw [->] (MEAS1.east) -- (RESET1.west);
	\node at ($(MEAS1.east) +(+0.8cm,+0.3cm)$) {\nm{\xvecest^+_{{\sss PRE},n}}};
	\node at ($(MEAS1.east) +(+0.8cm,-0.3cm)$) {\nm{\Pvec^+_{{\sss PRE},n}}};
	\draw [->] (RESET1.east) -- (VIS.west);
	\node at ($(RESET1.east) +(+0.8cm,+0.3cm)$) {\nm{\xvecest_{{\sss PRE},n}}};
	\node at ($(RESET1.east) +(+0.8cm,-0.3cm)$) {\nm{\vec x_{{\sss EST,PRE},n}}};
	
	\draw [->] ($(MEAS1.north) + (-1.5cm,+0.30cm)$) -| (MEAS1.north);
	\node at ($(MEAS1.north) +(-2.2cm,+0.30cm)$) {\nm{\xvectilde_s \setminus \vec I_i}};	
		
	\draw [->] ($(VIS.north) + (-1.5cm,+0.30cm)$) -| (VIS.north);
	\node at ($(VIS.north) +(-2.2cm,+0.30cm)$) {\nm{\vec I_i \subset \xvectilde_s}};	
	
	\node [block, below of=MEAS1, minimum width=2.2cm, node distance=2.5cm, align=center, minimum height=1.5cm] (MEAS2)  {\texttt{VA-INS} \\ \texttt{Measure Eq.}};
	\node [block, below of=RESET1, minimum width=2.2cm, node distance=2.5cm, align=center, minimum height=1.5cm] (RESET2) {\texttt{VA-INS} \\ \texttt{Reset Eq.}};
		
	\coordinate (turnpoint) at ($(MEAS2.north) + (-1.75cm,+0.50cm)$);
	\draw [->] (VIS.south) |- (turnpoint) |- ($(MEAS2.west)+(+0.0cm,+0.3cm)$);
	\node at ($(MEAS2.north) +(+3.0cm,+0.75cm)$){\nm{\xvecvis_i = \vec x_{{\sss IMG}i}}};
			
	\coordinate (divpoint) at ($(TIME.east) + (+0.4cm,+0.0cm)$);
	\filldraw [black] (divpoint) circle [radius=1pt];		
	\draw [->] (divpoint) |- ($(MEAS2.west)+(+0.0cm,-0.3cm)$);			
						
	\draw [->] (MEAS2.east) -- (RESET2.west);
	\node at ($(MEAS2.east) +(+0.8cm,+0.3cm)$) {\nm{\xvecest^+_n}};
	\node at ($(MEAS2.east) +(+0.8cm,-0.3cm)$) {\nm{\Pvec^+_n}};		
	\draw [->] (RESET2.east) -- node[pos=0.65] {\nm{\xvecest_n = \vec x_{{\sss EST}n}}} ($(RESET2.east)+(+3.5cm,+0.0cm)$);
				
	\coordinate (ResetOut) at ($(RESET2.south) + (-2.6cm,-0.40cm)$);
	\coordinate (TimeIn) at ($(TIME.south)  + (+2.6cm,-2.90cm)$);
	\draw [->] (RESET2.south) |- (ResetOut);
	\draw [dotted] (ResetOut) -- (TimeIn);
	\draw [->] (TimeIn) -| (TIME.south);
	\node at ($(RESET2.south) +(-1.8cm,-0.15cm)$) {\nm{\xvecest_n, \, \Pvec_n}};
	\node at ($(TIME.south) +(+1.2cm,-2.65cm)$) {\nm{\xvecest_{n-1}, \, \Pvec_{n-1}}};		
	
	\node at ($(VINS.west) +(+3.15cm,+2.55cm)$) {\texttt{VISUAL INERTIAL NAVIGATION SYSTEM}};
	\node at ($(VINS.east) +(-3.90cm,+2.55cm)$) {\nm{t_n = n \cdot \DeltatEST = t_s = s \cdot \DeltatSENSED = t_i = t \cdot \DeltatIMG}};
\end{tikzpicture}
\caption{\texttt{VINS} flow diagram when \nm{t_n = t_s = t_i}}
\label{fig:VINS_flow_diagram_times_equal}
\end{figure}

\item \texttt{GNSS}-Denied executions for which the \texttt{VVS} position and ground velocity observations are available at the required time (\nm{t_n = n \cdot \DeltatEST = t_i = i \cdot \DeltatIMG}) instead rely on the more complex scheme depicted in figure \ref{fig:VINS_flow_diagram_times_equal}. Note that this case only occurs in one out of ten executions for \texttt{GNSS}-Denied conditions. The first part of the process is exactly the same as in the previous case, and results in a preliminary estimated state that does not employ the \texttt{VVS} measurements (this is, does not employ the \nm{\vec I_i} image corresponding to \nm{t_i} nor the \texttt{IA-VNS} algorithms), denoted \nm{\hat{\vec x}_{\sss PRE}\lrp{t_n} = \vec x_{\sss EST,PRE}\lrp{t_n}} in the figure to avoid confusion.

This preliminary estimation, together with the \nm{\vec I_i} image, is then passed to the \texttt{IA-VNS} as described in \cite{VNSE} to obtain a visual state \nm{\xvecvis\lrp{t_i} = \xIMG\lrp{t_i}}, which as explained in section \ref{sec:VVS} is equivalent to the \texttt{VVS} position and ground velocity outputs. Only then it is possible to apply the complete \texttt{VA-INS} navigation filter measurement update and reset equations for a second time to obtain the final state estimation \nm{\xvecest\lrp{t_n} = \xEST\lrp{t_n}}.
\begin{figure}[h]
\centering
\begin{tikzpicture}[auto, node distance=2cm,>=latex']
	\node [coordinate](midinput) {};
	
	\coordinate (center) at ($(midinput) + (+0.0cm,+0.3cm)$);
	\pgfdeclarelayer{background};
	\node [rectangle, draw=black!50, fill=black!8, dashed, rounded corners, right of=center, node distance=5.0cm, minimum width=15.8cm, minimum height=3.3cm] (VINS) {};
	\pgfdeclarelayer{foreground};

	\node [block, right of=midinput, minimum width=2.2cm, node distance=0.0cm, align=center, minimum height=1.5cm] (TIME)  {\texttt{VA-INS} \\ \texttt{Time Eq.}};
	\node [block, right of=TIME, minimum width=2.2cm, node distance=4.5cm, align=center, minimum height=1.5cm] (MEAS)  {\texttt{VA-INS} \\ \texttt{Measure Eq.}};
	\node [block, right of=MEAS, minimum width=2.2cm, node distance=4.5cm, align=center, minimum height=1.5cm] (RESET) {\texttt{VA-INS} \\ \texttt{Reset Eq.}};
	
	\draw [->] ($(TIME.west)+(-1.5cm,+0.0cm)$) -- node[pos=0.25] {\nm{\xvecestzero}} (TIME.west);
	\draw [->] (TIME.east) -- node[pos=0.5] {\nm{\xvecest^-_n, \, \Pvec^-_n}} (MEAS.west);
	\draw [->] (MEAS.east) -- node[pos=0.5] {\nm{\xvecest^+_n, \, \Pvec^+_n}} (RESET.west);
	\draw [->] (RESET.east) -- node[pos=0.55] {\nm{\xvecest_n = \vec x_{{\sss EST}n}}} ($(RESET.east)+(+2.4cm,+0.0cm)$);
	
	\draw [->] ($(MEAS.north) + (-1.5cm,+0.30cm)$) -| (MEAS.north);
	\node at ($(MEAS.north) +(-2.2cm,+0.30cm)$) {\nm{\xvectilde_s \setminus \vec I_i}};	
		
	\coordinate (ResetOut) at ($(RESET.south) + (-3.4cm,-0.40cm)$);
	\coordinate (TimeIn) at ($(TIME.south)  + (+3.4cm,-0.40cm)$);
	
	\draw [->] (RESET.south) |- (ResetOut);
	\draw [dotted] (ResetOut) -- (TimeIn);
	\draw [->] (TimeIn) -| (TIME.south);
	
	\node at ($(RESET.south) +(-2.2cm,-0.15cm)$) {\nm{\xvecest_n, \, \Pvec_n}};
	\node at ($(TIME.south) +(+2.4cm,-0.15cm)$) {\nm{\xvecest_{n-1}, \, \Pvec_{n-1}}};		
	
	\node at ($(VINS.west) +(+3.15cm,+1.30cm)$) {\texttt{VISUAL INERTIAL NAVIGATION SYSTEM}};
	\node at ($(VINS.east) +(-4.10cm,+1.30cm)$) {\nm{t_n = n \cdot \DeltatEST = t_s = s \cdot \DeltatSENSED = t_g = g \cdot \DeltatGNSS}};
\end{tikzpicture}
\caption{\texttt{VINS} flow diagram when \nm{t_n = t_s = t_g}}
\label{fig:VINS_flow_diagram_times_equal_gnss}
\end{figure}

\item \texttt{GNSS}-Based executions for which the \texttt{GNSS} receiver position and ground velocity observations are available at the required time (\nm{t_n = n \cdot \DeltatEST = t_g = g \cdot \DeltatGNSS}). This case, which only occurs in one out of a hundred executions when \texttt{GNSS} signals are available and is shown in figure \ref{fig:VINS_flow_diagram_times_equal_gnss}, is in fact very similar to the first one above, with the only difference that the navigation filter observation system (\ref{eq:nav_vis_iner_filter_yn_so3_local}) can be applied as written since it includes the \texttt{GNSS} receiver observations.
\end{itemize}


\section{Testing: High Fidelity Simulation and Scenarios}\label{sec:Simulation}

To evaluate the performance of the proposed navigation algorithms, this article relies on Monte Carlo simulations consisting of one hundred runs each of two different scenarios based on the high fidelity stochastic flight simulator graphically depicted in figure \ref{fig:flow_diagram}. Described in detail in \cite{SIMULATION} and with its open source \nm{\CC} implementation available in \cite{CODE}, the simulator models the flight in varying weather and turbulent conditions of a fixed wing piston engine autonomous \texttt{UAV}.
\begin{figure}[h]
\centering
\begin{tikzpicture}[auto, node distance=2cm,>=latex']

	\node [coordinate](xrefinput) {};
	\node [blockgreen, right of=xrefinput, minimum width=2.6cm, node distance=3.0cm, align=center, minimum height=1.25cm] (GUIDANCE) {\texttt{GUIDANCE}};
	\draw [->] (xrefinput) -- node[pos=0.4] {\nm{\xREF}} (GUIDANCE.west);

	\node [blockgreen, right of=GUIDANCE, minimum width=2.6cm, node distance=5.0cm, align=center, minimum height=1.25cm] (CONTROL) {\texttt{CONTROL}};
	\draw [->] (GUIDANCE.east) -- node[pos=0.5] {\nm{\deltaTARGET}} (CONTROL.west);

	\node [blockyellow, right of=CONTROL, minimum width=1.8cm, node distance=3.5cm, align=center, minimum height=1.25cm] (AIRCRAFT) {\texttt{AIRCRAFT}};
	\draw [->] (CONTROL.east) --  node[pos=0.5] {\nm{\deltaCNTR}} (AIRCRAFT.west);

	\node [blockyellow, right of=AIRCRAFT, minimum width=1.8cm, node distance=2.6cm, align=center, minimum height=1.25cm] (EARTH) {\texttt{EARTH}};
	\draw [->] (EARTH.west) -- (AIRCRAFT.east);

	\node [coordinate, right of=AIRCRAFT, node distance=1.3cm] (midpoint){};
	\node [blockyellow, below of=midpoint, minimum width=1.8cm, node distance=2.5cm, align=center, minimum height=1.25cm] (FLIGHT) {\texttt{FLIGHT}};
	\draw [->] ($(EARTH.south)-(0.50cm,0cm)$) |- ($(FLIGHT.north)+(0.20cm,0.70cm)$) -- ($(FLIGHT.north)+(0.20cm,0cm)$);
	\draw [->] ($(AIRCRAFT.south)+(0.50cm,0cm)$) |- ($(FLIGHT.north)+(-0.20cm,0.70cm)$) -- ($(FLIGHT.north)-(0.20cm,0cm)$);

	\node [blockgreen, below of=GUIDANCE, minimum width=2.6cm, node distance=2.5cm, align=center, minimum height=1.25cm] (NAVIGATION) {\texttt{NAVIGATION}};
	\node [coordinate, left of=NAVIGATION, node distance=2.0cm] (pointnav1){};
	\node [coordinate, above of=pointnav1, node distance=1.2cm] (pointnav2){};
	\node [coordinate, below of=GUIDANCE, node distance=1.3cm] (pointnav3){};
	\filldraw [black] (pointnav3) circle [radius=1pt];
	\draw [->] ($(NAVIGATION.west)+(0cm,0.3cm)$) -| (pointnav2) -- (pointnav3) -- (GUIDANCE.south);
	\draw [->] (pointnav3) -| node[pos=0.25] {\nm{\xvecest = \xEST}} (CONTROL.south);
	\draw [->] ($(NAVIGATION.west)+(-1.5cm,-0.3cm)$) -- node[pos=0.4] {\nm{\xvecestzero}} ($(NAVIGATION.west)+(0cm,-0.3cm)$);

	\node [blockgreen, below of=CONTROL, minimum width=2.6cm, node distance=2.5cm, align=center, minimum height=1.25cm] (SENSORS) {\texttt{SENSORS}};
	\draw [->] (SENSORS.west) -- node[pos=0.5] {\nm{\xvectilde = \xSENSED}} (NAVIGATION.east);

	\node [coordinate, right of=EARTH, node distance=1.5cm] (pointflight1){};
	\node [coordinate, above of=EARTH, node distance=1.1cm] (pointflight2){};
	\filldraw [black] (pointflight2) circle [radius=1pt];
	\draw [->] (FLIGHT.west) -- node[pos=0.5] {\nm{\xvec = \xTRUTH}} (SENSORS.east);
	\draw [->] (FLIGHT.east) -| node[pos=0.25] {\nm{\xvec = \xTRUTH}} (pointflight1) |- (pointflight2) -| (AIRCRAFT.north);
	\draw [->] (pointflight2) -- (EARTH.north);
\end{tikzpicture}
\caption{Components of the high fidelity simulation}
\label{fig:flow_diagram}
\end{figure}

The simulator consists on two distinct processes. The first, represented by the yellow blocks on the right of figure \ref{fig:flow_diagram}, models the physics of flight and the interaction between the aircraft and its surroundings that results in the real aircraft trajectory \nm{\xvec = \xTRUTH}; the second, represented by the green blocks on the left, contains the aircraft systems in charge of ensuring that the resulting trajectory adheres as much as possible to the mission objectives. It includes the different sensors whose output comprise the sensed trajectory \nm{\xvectilde = \xSENSED}, the navigation system in charge of filtering it to obtain the estimated trajectory \nm{\xvecest = \xEST}, the guidance system that converts the reference objectives \nm{\xREF} into the control targets \nm{\deltaTARGET}, and the control system that adjusts the position of the throttle and aerodynamic control surfaces \nm{\deltaCNTR} so the estimated trajectory \nm{\xvecest} is as close as possible to the reference objectives \nm{\xREF}. Table \ref{tab:frequencies} lists the working frequencies of the various blocks represented in figure \ref{fig:flow_diagram}.

All components of the flight simulator have been modeled with as few simplifications as possible to increase the realism of the results, as explained in \cite{SENSORS,SIMULATION}. With the exception of the aircraft performances and its control system, which are deterministic, all other simulator components are treated as stochastic and hence vary from one execution to the next, enhancing the significance of the Monte Carlo simulation results. 

Most \texttt{VIO} packages discussed in appendix \ref{sec:GNSS-Denied} include in their release articles an evaluation when applied to the \texttt{EuRoC} Micro Air Vehicle (\texttt{MAV}) data sets\footnote{These data sets contain perfectly synchronized stereo images, \texttt{IMU} measurements, and ground truth readings obtained with laser, for eleven different indoor trajectories flown with a \texttt{MAV}, each with a duration in the order of two minutes and a total distance in the order of \nm{100 \, m}. This fact by itself indicates that the target application of exiting \texttt{VIO} implementations differs significantly from the main focus of this article, as there may exist accumulating errors that are completely non discernible after such short periods of time, but that grow non linearly and have the capability of inducing significant pose errors when the aircraft remains aloft for long periods of time.} \cite{Burri2016}, and so do independent articles such as \cite{Delmerico2018}. The algorithms presented in this article are hence tested through simulation under two different scenarios designed to analyze the consequences of losing the \texttt{GNSS} signals for long periods of time. Although a short summary is included below, detailed descriptions of the mission, weather, and wind field employed in each scenario can be found in \cite{SIMULATION}. Most parameters comprising the scenario are defined stochastically, resulting in different values for every execution. Note that all results shown in section \ref{sec:Results} are based on Monte Carlo simulations comprising one hundred runs of each scenario, testing the sensitivity of the proposed navigation algorithms to a wide variety of values in the parameters.
\begin{itemize}
\item Scenario \#1 has been defined with the objective of adequately representing the challenges faced by an autonomous fixed wing \texttt{UAV} that suddenly cannot rely on \texttt{GNSS} and hence changes course to reach a predefined recovery location situated at approximately one hour of flight time. In the process, in addition to executing an altitude and airspeed adjustment, the autonomous aircraft faces significant weather and wind field changes that make its \texttt{GNSS}-Denied navigation even more challenging. 

With respect to the mission, the stochastic parameters include the initial airspeed, pressure altitude, and bearing (\nm{\vtasINI, \HpINI, \chiINI}), their final values (\nm{\vtasEND, \HpEND, \chiEND}), and the time at which each of the three maneuvers is initiated\footnote{Turns are executed with a bank angle of \nm{\xiTURN = \pm 10 \, ^{\circ}}, altitude changes employ an aerodynamic path angle of \nm{\gammaTASCLIMB = \pm 2 \, ^{\circ}}, and airspeed modifications are automatically executed by the control system as set-point changes.}. The scenario lasts for \nm{\tEND = 3800 \, s}, while the \texttt{GNSS} signals are lost at \nm{\tGNSS = 100 \, s}.

The wind field is also defined stochastically, as its two parameters (speed and bearing) are constant both at the beginning (\nm{\vwindINI, \chiWINDINI}) and conclusion (\nm{\vwindEND, \chiWINDEND}) of the scenario, with a linear transition in between. The specific times at which the wind change starts and concludes also vary stochastically among the different simulation runs. As described in \cite{SIMULATION}, the turbulence remains strong throughout the whole scenario, but its specific values also vary stochastically from one execution to the next.

A similar linear transition occurs with the temperature and pressure offsets that define the atmospheric properties \cite{INSA}, as they are constant both at the start (\nm{\DeltaTINI, \DeltapINI}) and end (\nm{\DeltaTEND, \DeltapEND}) of the flight. In contrast with the wind field, the specific times at which the two transitions start and conclude are not only stochastic but also different from each other.

\item Scenario \#2 represents the challenges involved in continuing with the original mission upon the loss of the \texttt{GNSS} signals, executing a series of continuous turn maneuvers over a relatively short period of time with no atmospheric or wind variations. As in scenario \nm{\#1}, the \texttt{GNSS} signals are lost at \nm{\tGNSS = 100 \, s}, but the scenario duration is shorter (\nm{\tEND = 500 \, s}). The initial airspeed and pressure altitude (\nm{\vtasINI, \HpINI}) are defined stochastically and do not change throughout the whole scenario; the bearing however changes a total of eight times between its initial and final values, with all intermediate bearing values as well as the time for each turn varying stochastically from one execution to the next. Although the same turbulence is employed as in scenario \nm{\#1}, the wind and atmospheric parameters (\nm{\vwindINI, \chiWINDINI, \DeltaTINI, \DeltapINI}) remain constant throughout scenario \nm{\#2}.
\end{itemize}


\section{Navigation System Error in GNSS-Denied Conditions}\label{sec:Results}

This section presents the results obtained with the proposed fully integrated visual inertial navigation system (\texttt{VINS}) when executing Monte Carlo simulations of the two \texttt{GNSS}-Denied scenarios introduced in section \ref{sec:Simulation}, each consisting of one hundred executions. They are compared with those of the standalone inertial and visual systems (\texttt{INS} and \texttt{VNS}), as well as those of the inertially assisted \texttt{VNS} or \texttt{IA-VNS} described in section \ref{sec:architecture}. The tables below contain the \emph{navigation system error}\footnote{The \texttt{NSE} is the difference between the real states (\nm{\xvec}) and their estimation (\nm{\xvecest} or \nm{\xvecvis}) by the navigation system.} (\texttt{NSE}) incurred by the various navigation systems (and accordingly denoted as \texttt{INSE}, \texttt{VNSE}, \texttt{IA-VNSE}, and \texttt{VINSE}) at the conclusion of the two \texttt{GNSS}-Denied scenarios, represented by the mean, standard deviation, and maximum value of the estimation errors. In addition, the figures depict the variation with time of the \texttt{NSE} mean (solid line) and standard deviation (dashed lines) for the one hundred executions. The following remarks are necessary:
\begin{itemize}
\item The results obtained with the \texttt{INS} under the same two \texttt{GNSS}-Denied scenarios are discussed in \cite{INSE}. The attitude \texttt{INSE} does not drift and is bounded by the quality of the onboard sensors, ensuring the aircraft can remain aloft for as long as there is fuel available. The vertical position and ground velocity \texttt{INSE}s are also bounded by atmospheric physics and do not drift; their estimation errors depend on the atmospheric pressure offset and wind field changes that occur since the \texttt{GNSS} signals are lost. On the other hand, the horizontal position \texttt{INSE} drifts as a consequence of integrating the ground velocity without absolute observations. Of the six \nm{\mathbb{SE}(3)} degrees of freedom (three for attitude, two for horizontal position, one for altitude), the \texttt{INS} is hence capable of successfully estimating four of them in \texttt{GNSS}-Denied conditions.

\item Visual navigation systems (\texttt{VNS}, \texttt{IA-VNS}, and the fully integrated \texttt{VINS}) are only necessary to reduce the estimation error in the two remaining degrees of freedom (the horizontal position). Although they all estimate the complete six dimensional aircraft pose, their attitude and altitude estimations shall only be understood as a means to provide more accurate horizontal position estimations, which represents their sole objective.

\item The results obtained with the \texttt{VNS} and \texttt{IA-VNS} under the same two scenarios are discussed in \cite{VNSE}. Even though the \texttt{VNSE} slowly drifts in all six degrees of freedom, the addition of \texttt{INS} based priors enables the \texttt{IA-VNS} to reduce the drift in all six dimensions, with the resulting horizontal position \texttt{IA-VNSE} being just a fraction of the \texttt{INSE}. The attitude and altitude \texttt{IA-VNSE}s, although improved when compared to the \texttt{VNSE}s, are qualitatively inferior to the driftless \texttt{INSE}s.  

\item The main focus of this article lies in the horizontal position estimation improvements that occur when the navigation filter \texttt{EKF} can make use of the \texttt{VVS} observations provided by the \texttt{IA-VNS} to replace those of the \texttt{GNSS} receiver.
\end{itemize}


\subsection{Body Attitude Estimation}\label{subsec:Results_att}

Table \ref{tab:Res_euler} shows the \texttt{NSE} at the conclusion of both scenarios for the norm of the rotation vector between the real body attitude \nm{\qNB} and its estimations, \nm{\qNBest} or \nm{\qNBvis}. The errors hence can be formally written as \nm{\DeltarNBBestnorm = \|\qNBest \ominus \qNB\|} or \nm{\DeltarNBBvisnorm = \|\qNBvis \ominus \qNB\|}, where \nm{\ominus} represents the \nm{\mathbb{SO}\lrp{3}} minus operator \cite{LIE,Sola2018}. In addition, figures \ref{fig:Res_euler_base} and \ref{fig:Res_euler_alter} depict the variation with time of the body attitude \texttt{NSE} for both scenarios. 
\begin{center}
\begin{tabular}{llrrrr}
\hline
\multicolumn{2}{l}{Scenario \nm{\lrp{\tEND}}} & \multicolumn{1}{c}{\texttt{NSE}} & \multicolumn{1}{c}{\texttt{VNSE}} & \multicolumn{1}{c}{\texttt{IA-VNSE}} & \multicolumn{1}{c}{\texttt{VINSE}} \\
&\nm{\lrsb{^{\circ}}} & \nm{\DeltarNBBestnorm} & \nm{\DeltarNBBvisnorm} & \nm{\DeltarNBBvisnorm} & \nm{\DeltarNBBestnorm} \\
\hline
\multirow{3}{*}{\#1} & mean & \textbf{0.158} & \textbf{0.296} & \textbf{0.218} & \textbf{0.100} \\
                     & std  &         0.114  &         0.158  &         0.103  &         0.059  \\
                     & max  &         0.611  &         0.791  &         0.606  &         0.328  \\
\hline
\multirow{3}{*}{\#2} & mean & \textbf{0.128} & \textbf{0.253} & \textbf{0.221} & \textbf{0.107} \\
                     & std  &         0.078  &         0.161  &         0.137  &         0.068  \\
                     & max  &         0.369  &         0.730  &         0.788  &         0.377  \\
\hline
\end{tabular}
\end{center}
\captionof{table}{Aggregated final body attitude \texttt{INSE}, \texttt{VNSE}, \texttt{IA-VNSE}, and \texttt{VINSE} (100 runs)} \label{tab:Res_euler}

\begin{itemize}
\item After a short transition period following the introduction of \texttt{GNSS}-Denied conditions at \nm{\tGNSS = 100 \, s}, the body attitude \texttt{INSE} (blue lines) does not experience any drift with time in either scenario, and is bounded by the quality of the onboard sensors and the inertial navigation algorithms \cite{INSE}. The bounded attitude implies that the \texttt{UAV} can remain aloft without \texttt{GNSS} signals for as long as it has fuel.

\item The body attitude \texttt{VNSE} (red lines) experiences a slight drift with time in both scenarios, with most of the error attributed to the roll-in and roll-out maneuvers within the initial turn \cite{VNSE}. The lack of error growth during most of scenario \nm{\#2} is caused by the trajectories being so twisted (refer to \cite{SIMULATION}) that terrain zones previously mapped reappear in the camera field of view during the consecutive turns, and are hence employed by \texttt{SVO} as absolute references, resulting in a much better attitude estimation that what would occur under more spaced turns. 

\begin{figure}[h]
\centering
\pgfplotsset{
	every axis legend/.append style={
		at={(0.5,1.03)},
		anchor=south,
	},
}
\begin{tikzpicture}
\begin{axis}[
cycle list={{blue,no markers,very thick},
			{red,no markers,very thick},
			{orange,no markers,very thick},
			{green,no markers,very thick},
			{blue,dashed,no markers,ultra thin},{blue,dashed,no markers,ultra thin},
			{red,dashed,no markers,ultra thin},{red,dashed,no markers,ultra thin},
			{orange,dashed,no markers,ultra thin},{orange,dashed,no markers,ultra thin},
			{green,dashed,no markers,ultra thin},{green,dashed,no markers,ultra thin}},
width=16.0cm,
height=5.0cm,
xmin=0, xmax=3800, xtick={0,500,...,3500,3800},
xlabel={\nm{t \lrsb{s}}},
xmajorgrids,
ymin=0, ymax=0.4, ytick={0,0.1,0.2,0.3,0.4},
ylabel={\nm{\DeltarNBBestnorm, \, \DeltarNBBvisnorm \, \lrsb{^{\circ}}}},
ymajorgrids,
axis lines=left,
axis line style={-stealth},
legend entries={\nm{\mun{\DeltarNBBestnorm} \pm \sigman{\DeltarNBBestnorm}} \texttt{INSE},
                \nm{\mui{\DeltarNBBvisnorm} \pm \sigmai{\DeltarNBBvisnorm}} \texttt{VNSE},
				\nm{\mui{\DeltarNBBvisnorm} \pm \sigmai{\DeltarNBBvisnorm}} \texttt{IA-VNSE},
				\nm{\mun{\DeltarNBBestnorm} \pm \sigman{\DeltarNBBestnorm}} \texttt{VINSE}},
legend columns=2,
legend style={font=\footnotesize},
legend cell align=left,
]
\pgfplotstableread{figs/att/error_base_euler_deg_ins.txt}\mytablenav
\pgfplotstableread{figs/att/error_base_euler_deg_vns.txt}\mytablevis
\pgfplotstableread{figs/att/error_base_euler_deg_iavns.txt}\mytableiavis
\pgfplotstableread{figs/att/error_base_euler_deg_vins.txt}\mytablevins
\addplot table [header=false, x index=0,y index=1] {\mytablenav};
\addplot table [header=false, x index=0,y index=1] {\mytablevis};
\addplot table [header=false, x index=0,y index=1] {\mytableiavis};
\addplot table [header=false, x index=0,y index=1] {\mytablevins};
\addplot table [header=false, x index=0,y index=2] {\mytablenav};
\addplot table [header=false, x index=0,y index=3] {\mytablenav};
\addplot table [header=false, x index=0,y index=2] {\mytablevis};
\addplot table [header=false, x index=0,y index=3] {\mytablevis};
\addplot table [header=false, x index=0,y index=2] {\mytableiavis};
\addplot table [header=false, x index=0,y index=3] {\mytableiavis};
\addplot table [header=false, x index=0,y index=2] {\mytablevins};
\addplot table [header=false, x index=0,y index=3] {\mytablevins};
\path node [draw, shape=rectangle, fill=white] at (3000,0.35) {\footnotesize Scenario \nm{\#1}};
\end{axis}   
\end{tikzpicture}
\caption{Body attitude \texttt{INSE}, \texttt{VNSE}, \texttt{IA-VNSE}, and \texttt{VINSE} for scenario \nm{\#1} (100 runs)}
\label{fig:Res_euler_base}
\end{figure}

\item The \texttt{IA-VNSE} (orange lines) shows significant improvements over the \texttt{VNSE} as the \texttt{IA-VNS} successfully limits the deviation between the visual pitch and bank estimations with respect to the inertial ones provided by the \texttt{INS} (figure \ref{fig:flow_diagram_iavns}) \cite{VNSE}. The improvement is more significant in the case of scenario \nm{\#1} because the \texttt{IA-VNS} algorithms are by design slow adjustments that require significant time to slowly correct the deviations \cite{VNSE}. Note that the higher \texttt{IA-VNSE}s at the beginning of both scenarios are also caused by the \texttt{IA-VNS} initialization adjustments \cite{VNSE}.

\begin{figure}[h]
\centering
\pgfplotsset{
	every axis legend/.append style={
		at={(0.50,1.03)},
		anchor=south,
	},
}
\begin{tikzpicture}
\begin{axis}[
cycle list={{blue,no markers,very thick},
            {red,no markers,very thick},
			{orange,no markers,very thick},
			{green,no markers,very thick},
            {blue,dashed,no markers,ultra thin},{blue,dashed,no markers,ultra thin},
            {red,dashed,no markers,ultra thin},{red,dashed,no markers,ultra thin},
			{orange,dashed,no markers,ultra thin},{orange,dashed,no markers,ultra thin},
			{green,dashed,no markers,ultra thin},{green,dashed,no markers,ultra thin}},
width=14.0cm,
height=5.0cm,
xmin=0, xmax=500, xtick={0,50,...,500},
xlabel={\nm{t \lrsb{s}}},
xmajorgrids,
ymin=0, ymax=0.4, ytick={0,0.1,0.2,0.3,0.4},
ylabel={\nm{\DeltarNBBestnorm, \, \DeltarNBBvisnorm \, \lrsb{^{\circ}}}},
ymajorgrids,
axis lines=left,
axis line style={-stealth},			
legend entries={\nm{\mun{\DeltarNBBestnorm} \pm \sigman{\DeltarNBBestnorm}} \texttt{INSE},
                \nm{\mui{\DeltarNBBvisnorm} \pm \sigmai{\DeltarNBBvisnorm}} \texttt{VNSE},
				\nm{\mui{\DeltarNBBvisnorm} \pm \sigmai{\DeltarNBBvisnorm}} \texttt{IA-VNSE},
				\nm{\mun{\DeltarNBBestnorm} \pm \sigman{\DeltarNBBestnorm}} \texttt{VINSE}},
legend columns=2,
legend style={font=\footnotesize},
legend cell align=left,
]
\pgfplotstableread{figs/att/error_alter_euler_deg_ins.txt}\mytablenav
\pgfplotstableread{figs/att/error_alter_euler_deg_vns.txt}\mytablevis
\pgfplotstableread{figs/att/error_alter_euler_deg_iavns.txt}\mytableiavis
\pgfplotstableread{figs/att/error_alter_euler_deg_vins.txt}\mytablevins
\addplot table [header=false, x index=0,y index=1] {\mytablenav};
\addplot table [header=false, x index=0,y index=1] {\mytablevis};
\addplot table [header=false, x index=0,y index=1] {\mytableiavis};
\addplot table [header=false, x index=0,y index=1] {\mytablevins};
\addplot table [header=false, x index=0,y index=2] {\mytablenav};
\addplot table [header=false, x index=0,y index=3] {\mytablenav};
\addplot table [header=false, x index=0,y index=2] {\mytablevis};
\addplot table [header=false, x index=0,y index=3] {\mytablevis};
\addplot table [header=false, x index=0,y index=2] {\mytableiavis};
\addplot table [header=false, x index=0,y index=3] {\mytableiavis};
\addplot table [header=false, x index=0,y index=2] {\mytablevins};
\addplot table [header=false, x index=0,y index=3] {\mytablevins};
\path node [draw, shape=rectangle, fill=white] at (75,0.30) {\footnotesize Scenario \nm{\#2}};
\end{axis}   
\end{tikzpicture}
\caption{Body attitude \texttt{INSE}, \texttt{VNSE}, \texttt{IA-VNSE}, and \texttt{VINSE} for scenario \nm{\#2} (100 runs)}
\label{fig:Res_euler_alter}
\end{figure}

\item The fully integrated \texttt{VINSE} shows sustained and significant body attitude estimation improvements with respect to the \texttt{INSE} in both scenarios, which can be attributed to various factors. On one hand, the proposed navigation filter successfully merges the incremental horizontal displacement observations supplied by the \texttt{VVS} with the bearing provided by the magnetometers, resulting in better yaw estimations. In addition, the absence of simplifications in the \texttt{EKF} observation equations\footnote{Note that in the absence of both \texttt{GNSS} receiver and \texttt{VVS} observations, the filter observation equations need to be simplified to protect the filter behavior from the negative effects of inaccurate velocity and position estimations, as discussed in \cite{INSE}.} ensures that the benefits also reach the already accurate inertial pitch and bank angle estimations. Additional body attitude estimation improvements are achieved by closing the loop (figure \ref{fig:flow_diagram_vins}) and directly employing the increasingly accurate body attitude inertial \texttt{VA-INS} estimations to feed the \texttt{IA-VNS} (section \ref{sec:VINS}).
\end{itemize}


\subsection{Vertical Position Estimation}\label{subsec:Results_h}

Table \ref{tab:Res_h} contains the vertical position \texttt{NSE} (\nm{\Delta\hest = \hest - h}, \nm{\Delta\hvis = \hvis - h}) at the conclusion of both scenarios, which can be considered unbiased or zero mean in all eight cases (two scenarios and four estimation methods) as the mean is always significantly lower than both the standard deviation or the maximum error value. The geometric altitude \texttt{NSE} evolution with time is depicted in figures \ref{fig:Res_h_base} and \ref{fig:Res_h_alter}, respectively. 
\begin{center}
\begin{tabular}{llrrrr}
\hline
\multicolumn{2}{l}{Scenario \nm{\lrp{\tEND}}} & \multicolumn{1}{c}{\texttt{INSE}} & \multicolumn{1}{c}{\texttt{VNSE}} & \multicolumn{1}{c}{\texttt{IA-VNSE}} & \multicolumn{1}{c}{\texttt{VINSE}} \\ 
& \nm{\lrsb{m}} & \multicolumn{1}{c}{\nm{\Delta\hest}} & \multicolumn{1}{c}{\nm{\Delta\hvis}} & \multicolumn{1}{c}{\nm{\Delta\hvis}} & \multicolumn{1}{c}{\nm{\Delta\hest}}  \\
\hline
\multirow{3}{*}{\#1} & mean &          -4.18  &          +82.91  &          +22.86  &          -3.97  \\
                     & std  & \textbf{ 25.78} & \textbf{ 287.58} & \textbf{  49.17} & \textbf{ 26.12} \\
                     & max  &         -70.49  &         +838.32  &         +175.76  &         -70.47  \\
\hline
\multirow{3}{*}{\#2} & mean &          +0.76  &          +3.45  &          +3.59  &          +0.74  \\
                     & std  & \textbf{  7.55} & \textbf{ 20.56} & \textbf{ 13.01} & \textbf{  7.60} \\
                     & max  &         -19.86  &         +72.69  &         +71.64  &         -18.86  \\
\hline
\end{tabular}
\end{center}
\captionof{table}{Aggregated final vertical position \texttt{INSE}, \texttt{VNSE}, \texttt{IA-VNSE}, and \texttt{VINSE} (100 runs)} \label{tab:Res_h}

\begin{itemize}
\item The geometric altitude \texttt{INSE} (blue lines) is bounded by the change in atmospheric pressure offset \cite{INSA} since the time the \texttt{GNSS} signals are lost \cite{INSE}.

\item The vertical position \texttt{VNSE} (red lines) is significantly worse than the \texttt{INSE} both qualitatively and quantitatively \cite{VNSE}, experiencing a continuous drift or error growth with time; the errors are logically bigger for scenario \nm{\#1} because of its much longer duration. Most of this drift results from adding the estimated relative pose between two consecutive images to a pose (that of the previous image) with an attitude that already possesses a small pitch error (refer to the attitude estimation analysis in section \ref{subsec:Results_att}). Note that even a fraction of a degree deviation in pitch can result in hundreds of meters in vertical error when applied to the total distance flown in scenario \nm{\#1}. As shown in figure \ref{fig:Res_h_alter}, the vertical position \texttt{VNSE} grows more slowly in the second half of scenario \nm{\#2} because, as explained in section \ref{subsec:Results_att} above, continuous turn maneuvers cause previously mapped terrain points to reappear in the camera field of view, stopping the growth in the attitude error (pitch included), which indirectly has the effect of slowing the growth in altitude estimation error.

\begin{figure}[h]
\centering
\pgfplotsset{
	every axis legend/.append style={
		at={(0.50,1.03)},
		anchor=south,
	},
}
\begin{tikzpicture}
\begin{axis}[
cycle list={{blue,no markers,very thick},
            {red,no markers,very thick},
			{orange,no markers,very thick},
			{green,no markers,very thick},
            {blue,dashed,no markers},{blue,dashed,no markers},
			{red,dashed,no markers},{red,dashed,no markers},
			{orange,dashed,no markers},{orange,dashed,no markers},
			{green,dashed,no markers},{green,dashed,no markers}},
width=16.0cm,
height=5.0cm,
xmin=0, xmax=3800, xtick={0,500,...,3500,3800},
xlabel={\nm{t \lrsb{s}}},
xmajorgrids,
ymin=-100, ymax=100, ytick={-100,-50,...,100},
ylabel={\nm{\Delta\hest, \, \Delta\hvis \, \lrsb{m}}},
ymajorgrids,
axis lines=left,
axis line style={-stealth},
legend entries={\nm{\mun{\Delta\hest} \pm \sigman{\Delta\hest}} \texttt{INSE},
                \nm{\mui{\Delta\hvis} \pm \sigmai{\Delta\hvis}} \texttt{VNSE},
				\nm{\mui{\Delta\hvis} \pm \sigmai{\Delta\hvis}} \texttt{IA-VNSE},
				\nm{\mun{\Delta\hest} \pm \sigman{\Delta\hest}} \texttt{VINSE}},
legend columns=4,
legend style={font=\footnotesize},
legend cell align=left,
]
\draw [] (0.0,0.0) -- (3800.0,0.0);
\pgfplotstableread{figs/h/error_base_h_m_ins.txt}\mytablenav
\pgfplotstableread{figs/h/error_base_h_m_vns.txt}\mytablevis
\pgfplotstableread{figs/h/error_base_h_m_iavns.txt}\mytableiavis
\pgfplotstableread{figs/h/error_base_h_m_vins.txt}\mytablevins
\addplot table [header=false, x index=0,y index=1] {\mytablenav};
\addplot table [header=false, x index=0,y index=1] {\mytablevis};
\addplot table [header=false, x index=0,y index=1] {\mytableiavis};
\addplot table [header=false, x index=0,y index=1] {\mytablevins};
\addplot table [header=false, x index=0,y index=2] {\mytablenav};
\addplot table [header=false, x index=0,y index=3] {\mytablenav};
\addplot table [header=false, x index=0,y index=2] {\mytablevis};
\addplot table [header=false, x index=0,y index=3] {\mytablevis};
\addplot table [header=false, x index=0,y index=2] {\mytableiavis};
\addplot table [header=false, x index=0,y index=3] {\mytableiavis};
\addplot table [header=false, x index=0,y index=2] {\mytablevins};
\addplot table [header=false, x index=0,y index=3] {\mytablevins};
\path node [draw, shape=rectangle] at (400,75) {\footnotesize Scenario \nm{\#1}};
\path node [draw, shape=rectangle] at (2600,-75) {\footnotesize \textcolor{green}{\nm{\mu_{VINSE}, \sigma_{VINSE}}} overlaps \textcolor{blue}{\nm{\mu_{INSE}, \sigma_{INSE}}}};
\end{axis}   
\end{tikzpicture}
\caption{Vertical position \texttt{INSE}, \texttt{VNSE}, \texttt{IA-VNSE}, and \texttt{VINSE} for scenario \nm{\#1} (100 runs)} \label{fig:Res_h_base}
\end{figure}

\item The \texttt{IA-VNSE} (orange lines) experiences a drastic reduction with respect to the \texttt{VNSE} in the case of scenario \nm{\#1}, where its extended duration allows the \texttt{IA-VNS} slow pose adjustments to accumulate into significant altitude corrections over time, and less pronounced but nevertheless significant for scenario \nm{\#2} \cite{VNSE}. Although it can not be compared with the vertical position results obtained with the \texttt{INS}, note that its main purpose is to improve the fit between the real terrain and the map generated by \texttt{SVO}, resulting in horizontal position improvements (section \ref{subsec:Results_hor}).

\begin{figure}[h]
\centering
\pgfplotsset{
	every axis legend/.append style={
		at={(0.50,1.03)},
		anchor=south,
	},
}
\begin{tikzpicture}
\begin{axis}[
cycle list={{blue,no markers,very thick},
            {red,no markers,very thick},
			{orange,no markers,very thick},
			{green,no markers,very thick},
            {blue,dashed,no markers},{blue,dashed,no markers},
			{red,dashed,no markers},{red,dashed,no markers},
			{orange,dashed,no markers},{orange,dashed,no markers},
			{green,dashed,no markers},{green,dashed,no markers}},
width=14.0cm,
height=5.0cm,
xmin=0, xmax=500, xtick={0,50,...,500},
xlabel={\nm{t \lrsb{s}}},
xmajorgrids,
ymin=-10, ymax=15, ytick={-10,-5,0,5,10,15},
ylabel={\nm{\Delta\hest, \, \Delta\hvis \, \lrsb{m}}},
ymajorgrids,
axis lines=left,
axis line style={-stealth},
legend entries={\nm{\mun{\Delta\hest} \pm \sigman{\Delta\hest}} \texttt{INSE},
                \nm{\mui{\Delta\hvis} \pm \sigmai{\Delta\hvis}} \texttt{VNSE},
				\nm{\mui{\Delta\hvis} \pm \sigmai{\Delta\hvis}} \texttt{IA-VNSE},
				\nm{\mun{\Delta\hest} \pm \sigman{\Delta\hest}} \texttt{VINSE}}, 
legend columns=4,
legend style={font=\footnotesize},
legend cell align=left,
]
\draw [] (0.0,0.0) -- (500.0,0.0);
\pgfplotstableread{figs/h/error_alter_h_m_ins.txt}\mytablenav
\pgfplotstableread{figs/h/error_alter_h_m_vns.txt}\mytablevis
\pgfplotstableread{figs/h/error_alter_h_m_iavns.txt}\mytableiavis
\pgfplotstableread{figs/h/error_alter_h_m_vins.txt}\mytablevins
\addplot table [header=false, x index=0,y index=1] {\mytablenav};
\addplot table [header=false, x index=0,y index=1] {\mytablevis};
\addplot table [header=false, x index=0,y index=1] {\mytableiavis};
\addplot table [header=false, x index=0,y index=1] {\mytablevins};
\addplot table [header=false, x index=0,y index=2] {\mytablenav};
\addplot table [header=false, x index=0,y index=3] {\mytablenav};
\addplot table [header=false, x index=0,y index=2] {\mytablevis};
\addplot table [header=false, x index=0,y index=3] {\mytablevis};
\addplot table [header=false, x index=0,y index=2] {\mytableiavis};
\addplot table [header=false, x index=0,y index=3] {\mytableiavis};
\addplot table [header=false, x index=0,y index=2] {\mytablevins};
\addplot table [header=false, x index=0,y index=3] {\mytablevins};
\path node [draw, shape=rectangle] at (60,-3.0) {\footnotesize Scenario \nm{\#2}};
\path node [draw, shape=rectangle] at (315,-3.0) {\footnotesize \textcolor{green}{\nm{\mu_{VINSE}, \sigma_{VINSE}}} overlaps \textcolor{blue}{\nm{\mu_{INSE}, \sigma_{INSE}}}};
\end{axis}   
\end{tikzpicture}
\caption{Vertical position \texttt{INSE}, \texttt{VNSE}, \texttt{IA-VNSE}, and \texttt{VINSE} for scenario \nm{\#2} (100 runs)}
\label{fig:Res_h_alter}
\end{figure}

\item The fully integrated \texttt{VINS} (green lines) vertical position estimation accuracy is virtually the same as that of the \texttt{INS} because the proposed navigation filter described in section \ref{sec:VA-INS} relies on freezing the pressure offset estimation \nm{\Deltapest} when the \texttt{GNSS} signals are lost, exactly the same as the \texttt{INS} filter \cite{INSE}. The resulting altitude estimations are unbiased and bounded, with an error that depends on the change in pressure offset since the time the \texttt{GNSS} are lost.
\end{itemize}


\subsection{Horizontal Position Estimation}\label{subsec:Results_hor}

Table \ref{tab:Res_hor} lists the mean, standard deviation, and maximum value of the horizontal position \texttt{NSE}s at the conclusion of the two scenarios, both in length units and as a percentage of the total distance flown in \texttt{GNSS}-Denied conditions. The horizontal position estimation capabilities of the four considered navigation systems share the fact that all of them exhibit an unrestrained drift or growth with time, as shown in figures \ref{fig:Res_hor_base} and \ref{fig:Res_hor_alter}. 
\begin{center}
\begin{tabular}{llrrrrrrrrr}
\hline
\multicolumn{3}{l}{Scenario \nm{\lrp{\tEND}}} & \multicolumn{2}{c}{\texttt{INSE}} & \multicolumn{2}{c}{\texttt{VNSE}} & \multicolumn{2}{c}{\texttt{IA-VNSE}} & \multicolumn{2}{c}{\texttt{VINSE}} \\
& & \multicolumn{1}{c}{Distance} & \multicolumn{2}{c}{\nm{\Delta \hat{x}_{\sss HOR}}} & \multicolumn{2}{c}{\nm{\Delta \circled{x}_{\sss HOR}}} & \multicolumn{2}{c}{\nm{\Delta \circled{x}_{\sss HOR}}} & \multicolumn{2}{c}{\nm{\Delta \hat{x}_{\sss HOR}}} \\
& & \multicolumn{1}{c}{\nm{\lrsb{m}}} & \multicolumn{1}{c}{\nm{\lrsb{m}}} & \multicolumn{1}{c}{\nm{\lrsb{\%}}} & \multicolumn{1}{c}{\nm{\lrsb{m}}} & \multicolumn{1}{c}{\nm{\lrsb{\%}}} & \multicolumn{1}{c}{\nm{\lrsb{m}}} & \multicolumn{1}{c}{\nm{\lrsb{\%}}} & \multicolumn{1}{c}{\nm{\lrsb{m}}} & \multicolumn{1}{c}{\nm{\lrsb{\%}}} \\
\hline
\multirow{3}{*}{\nm{\#1}} & mean & 107873 &  7276 & \textbf{ 7.10} &  4179 &  \textbf{ 3.82} &  488 & \textbf{0.46} &  207 & \textbf{0.19} \\
                          & std  &  19756 &  4880 &          5.69  &  3308 &           2.73  &  350 &         0.31  &  185 &         0.15  \\
                          & max  & 172842 & 25288 &         32.38  & 21924 &          14.22  & 1957 &         1.48  & 1257 &         1.09  \\
\hline
\multirow{3}{*}{\nm{\#2}} & mean & 14198 & 216 & \textbf{1.52} & 251 & \textbf{1.77} &  33 & \textbf{0.23} & 18 & \textbf{0.13} \\
                          & std  &  1176 & 119 &         0.86  & 210 &         1.48  &  26 &         0.18  &  9 &         0.07  \\
                          & max  & 18253 & 586 &         4.38  & 954 &         7.08  & 130 &         0.98  & 43 &         0.33  \\
\hline
\end{tabular}
\end{center}
\captionof{table}{Aggregated final horizontal position \texttt{INSE}, \texttt{VNSE}, \texttt{IA-VNSE}, and \texttt{VINSE} (100 runs)} \label{tab:Res_hor}

\begin{itemize}
\item In the case of the standalone \texttt{INS} (blue lines), the drift is caused by integrating the bounded horizontal velocity estimations without any absolute observations (which in normal circumstances are provided by the \texttt{GNSS} receiver) \cite{INSE}, resulting in the aircraft incurring in position errors that preclude inertial systems as a valid navigation source for \texttt{GNSS}-Denied navigation. Note that as explained in \cite{INSE}, the errors (percentage wise) are much lower in scenario \nm{\#2} because it does not include any wind variations, resulting in a more accurate ground velocity (and hence horizontal position) estimation than what could be achieved otherwise.

\item The horizontal position \texttt{VNSE} (red lines) also shows a continuous error accumulation caused by the concatenation of the relative poses between consecutive images without absolute references, coupled with additional inaccuracies traced to the estimation of the aircraft height over the terrain during the visual odometry initialization \cite{VNSE}. Figure \ref{fig:Res_hor_alter} shows the same slope diminution in the second half of the scenario discussed in previous sections caused by previously mapped terrain points reappearing in the camera field of view as a consequence of the continuous turns present in scenario \nm{\#2}.

\item The \texttt{IA-VNS} (orange lines) results in major horizontal position estimation improvements over the \texttt{VNS} in both scenarios, as the mean of the final errors diminishes from \nm{3.82} to \nm{0.46 \, \%} for scenario \nm{\#1}, and from \nm{1.77} to \nm{0.23 \, \%} for scenario \nm{\#2}. The repeatability of the results also improves, as the standard deviation falls from \nm{2.73} to \nm{0.31 \, \%} and from \nm{1.48} to \nm{0.18 \, \%} for both scenarios. As explained in \cite{VNSE}, the reason for this improvement lies in that the improved \texttt{IA-VNS} attitude and altitude estimations discussed in sections \ref{subsec:Results_att} and \ref{subsec:Results_h} enable \texttt{SVO} to build a better map of the terrain, resulting in an improved fit between the ground terrain and associated 3D points depicted in the images on one side, and the estimated aircraft pose indicating the position and attitude from where the images are taken on the other.

\begin{figure}[h]
\centering
\pgfplotsset{
	every axis legend/.append style={
		at={(0.5,1.03)},
		anchor=south,
	},
}
\begin{tikzpicture}
\begin{axis}[
cycle list={{blue,no markers,very thick},
			{red,no markers,very thick},
			{orange,no markers,very thick},
			{green,no markers,very thick},
			{blue,dashed,no markers,ultra thin},{blue,dashed,no markers,ultra thin},
			{red,dashed,no markers,ultra thin},{red,dashed,no markers,ultra thin},
			{orange,dashed,no markers,ultra thin},{orange,dashed,no markers,ultra thin},
			{green,dashed,no markers,ultra thin},{green,dashed,no markers,ultra thin}},			
width=16.0cm,
height=5.0cm,
xmin=0, xmax=3800, xtick={0,500,...,3500,3800},
xlabel={\nm{t \lrsb{s}}},
xmajorgrids,
ymin=0, ymax=4000, ytick={0,1000,...,4000},
ylabel={\nm{\Deltaxhorest, \, \Deltaxhorvis \, \lrsb{m}}},
ymajorgrids,
axis lines=left,
axis line style={-stealth},
legend entries={\nm{\mun{\Deltaxhorest} \pm \sigman{\Deltaxhorest}} \texttt{INSE},
				\nm{\mui{\Deltaxhorvis} \pm \sigmai{\Deltaxhorvis}} \texttt{VNSE},
				\nm{\mui{\Deltaxhorvis} \pm \sigmai{\Deltaxhorvis}} \texttt{IA-VNSE},
				\nm{\mun{\Deltaxhorest} \pm \sigman{\Deltaxhorest}} \texttt{VINSE}},
legend columns=2,
legend style={font=\footnotesize},
legend cell align=left,
]
\pgfplotstableread{figs/hor/error_base_hor_m_pc_ins.txt}\mytableins
\pgfplotstableread{figs/hor/error_base_hor_m_pc_vns.txt}\mytablevns
\pgfplotstableread{figs/hor/error_base_hor_m_pc_iavns.txt}\mytableiavns
\pgfplotstableread{figs/hor/error_base_hor_m_pc_vins.txt}\mytablevins
\addplot table [header=false, x index=0,y index=1] {\mytableins};
\addplot table [header=false, x index=0,y index=1] {\mytablevns};
\addplot table [header=false, x index=0,y index=1] {\mytableiavns};
\addplot table [header=false, x index=0,y index=1] {\mytablevins};
\addplot table [header=false, x index=0,y index=2] {\mytableins};
\addplot table [header=false, x index=0,y index=3] {\mytableins};
\addplot table [header=false, x index=0,y index=2] {\mytablevns};
\addplot table [header=false, x index=0,y index=3] {\mytablevns};
\addplot table [header=false, x index=0,y index=2] {\mytableiavns};
\addplot table [header=false, x index=0,y index=3] {\mytableiavns};
\addplot table [header=false, x index=0,y index=2] {\mytablevins};
\addplot table [header=false, x index=0,y index=3] {\mytablevins};
\path node [draw, shape=rectangle, fill=white] at (500,3000) {\footnotesize Scenario \nm{\#1}};
\end{axis}   
\end{tikzpicture}
\caption{Horizontal position \texttt{INSE}, \texttt{VNSE}, \texttt{IA-VNSE}, and \texttt{VINSE} for scenario \nm{\#1} (100 runs)}
\label{fig:Res_hor_base}
\end{figure}

\item In the case of the fully integrated \texttt{VINS} (green lines), the objective when supplying the proposed navigation filter with the \texttt{VVS} observations (equivalent to the \texttt{IA-VNS} incremental displacement estimations) is to employ more accurate observation equations within the \texttt{EKF} as there is no longer a need to protect the filter from the negative effects of inaccurate velocity and position values, as in the case of the \texttt{INS} filter \cite{INSE}. This improved design enables significant improvements in body attitude (section \ref{subsec:Results_att}) because the filter is capable of incorporating the whole \texttt{IA-VNS} horizontal position improvements into the \texttt{EKF} filter. In a positive feedback loop, the \texttt{VINS} then capitalizes on its more accurate attitude estimations to further improve its internal \texttt{IA-VNS} horizontal position estimations (figure \ref{fig:flow_diagram_vins}). Although small in absolute terms (length units) when compared to the much higher errors encountered in previous phases, percentage-wise the horizontal position improvements obtained when closing the loop are very significant, as the final error mean diminishes from \nm{0.46} to \nm{0.19 \, \%} for scenario \nm{\#1}, and from \nm{0.23} to \nm{0.13 \, \%} for \nm{\#2}, while the standard deviation also falls from \nm{0.31} to \nm{0.15 \, \%}, and from \nm{0.18} to \nm{0.07 \, \%}, respectively.
\end{itemize}

\begin{figure}[h]
\centering
\pgfplotsset{
	every axis legend/.append style={
		at={(0.50,1.05)},
		anchor=south,
	},
}
\begin{tikzpicture}
\begin{axis}[
cycle list={{blue,no markers,very thick},
			{red,no markers,very thick},
			{orange,no markers,very thick},
			{green,no markers,very thick},
			{blue,dashed,no markers,ultra thin},{blue,dashed,no markers,ultra thin},
			{red,dashed,no markers,ultra thin},{red,dashed,no markers,ultra thin},
			{orange,dashed,no markers,ultra thin},{orange,dashed,no markers,ultra thin},
			{green,dashed,no markers,ultra thin},{green,dashed,no markers,ultra thin}},
width=14.0cm,
height=5.0cm,
xmin=0, xmax=500, xtick={0,50,...,500},
xlabel={\nm{t \lrsb{s}}},
xmajorgrids,
ymin=0, ymax=250, ytick={0,50,...,250},
ylabel={\nm{\Deltaxhorest, \, \Deltaxhorvis \, \lrsb{m}}},
ymajorgrids,
axis lines=left,
axis line style={-stealth},
legend entries={\nm{\mun{\Deltaxhorest} \pm \sigman{\Deltaxhorest}} \texttt{INSE},
				\nm{\mui{\Deltaxhorvis} \pm \sigmai{\Deltaxhorvis}} \texttt{VNSE},
				\nm{\mui{\Deltaxhorvis} \pm \sigmai{\Deltaxhorvis}} \texttt{IA-VNSE},
				\nm{\mun{\Deltaxhorest} \pm \sigman{\Deltaxhorest}} \texttt{VINSE}},
legend columns=2,
legend style={font=\footnotesize},
legend cell align=left,
]
\pgfplotstableread{figs/hor/error_alter_hor_m_pc_ins.txt}\mytableins
\pgfplotstableread{figs/hor/error_alter_hor_m_pc_vns.txt}\mytablevns
\pgfplotstableread{figs/hor/error_alter_hor_m_pc_iavns.txt}\mytableiavns
\pgfplotstableread{figs/hor/error_alter_hor_m_pc_vins.txt}\mytablevins
\addplot table [header=false, x index=0,y index=1] {\mytableins};
\addplot table [header=false, x index=0,y index=1] {\mytablevns};
\addplot table [header=false, x index=0,y index=1] {\mytableiavns};
\addplot table [header=false, x index=0,y index=1] {\mytablevins};
\addplot table [header=false, x index=0,y index=2] {\mytableins};
\addplot table [header=false, x index=0,y index=3] {\mytableins};
\addplot table [header=false, x index=0,y index=2] {\mytablevns};
\addplot table [header=false, x index=0,y index=3] {\mytablevns};
\addplot table [header=false, x index=0,y index=2] {\mytableiavns};
\addplot table [header=false, x index=0,y index=3] {\mytableiavns};
\addplot table [header=false, x index=0,y index=2] {\mytablevins};
\addplot table [header=false, x index=0,y index=3] {\mytablevins};
\path node [draw, shape=rectangle, fill=white] at (50,200) {\footnotesize Scenario \nm{\#2}};
\end{axis}   
\end{tikzpicture}
\caption{Horizontal position \texttt{INSE}, \texttt{VNSE}, \texttt{IA-VNSE}, and \texttt{VINSE} for scenario \nm{\#2} (100 runs)}
\label{fig:Res_hor_alter}
\end{figure}

This two step reduction in the horizontal position error means that, although drift continues to be present due to the lack of absolute position observations in \texttt{GNSS}-Denied conditions, the horizontal position \texttt{VINSE}s are low enough so as to enable the autonomous aircraft to reach the vicinity of distant recovery locations in \texttt{GNSS}-Denied conditions, even when faced with turbulent and varying weather.


\section{Summary of Results} \label{sec:Summary}

This article proposes a \texttt{VINS} intended for fixed wing autonomous \texttt{UAV}s flying in \texttt{GNSS}-Denied conditions, which is composed by two algorithms (inertial and visual) that simultaneously control and assist each other. On one side, the \texttt{IA-VNS} presented in \cite{VNSE} complements a publicly available visual odometry pipeline such as \texttt{SVO} \cite{Forster2014, Forster2016} with priors based on the attitude and altitude \texttt{VA-INS} estimations so its visual estimations do not deviate in excess from its inertial counterparts. Its accurate incremental displacement outputs can be considered as the measurements of a virtual sensor (\texttt{VVS}) and employed by the \texttt{VA-INS} \texttt{EKF} as replacements for the \texttt{GNSS} receiver observations, as described in sections \ref{sec:VA-INS} and \ref{sec:VINS}, resulting in major improvements when estimating the aircraft position without the use of \texttt{GNSS} signals. The results obtained when applying the proposed algorithms to high fidelity Monte Carlo simulations of two scenarios representative of the challenges of \texttt{GNSS}-Denied navigation indicate the following:

\begin{itemize}
\item The \texttt{VINS} \textbf{body attitude} estimation is qualitatively similar to that obtained by the standalone \texttt{INS} \cite{INSE}, which relies exclusively on an inertial filter without any visual algorithms. The bounded estimations (unbiased or zero mean for each Euler angle, biased for the total error as it is a norm) enable the aircraft to remain aloft in \texttt{GNSS}-Denied conditions for as long as it has fuel. Quantitatively, the \texttt{VVS} observations and the associated more accurate filter equations result in significant body attitude accuracy improvements in the two considered scenarios.

\item The \texttt{VINS} \textbf{vertical position} estimation is qualitatively and quantitatively similar to that of the standalone \texttt{INS} \cite{INSE} as both share the same algorithm, which freezes the atmospheric pressure offset estimation at the time that the \texttt{GNSS} signals are lost. In addition to ionospheric effects (which also apply when \texttt{GNSS} signals are available), the altitude error depends on the amount of pressure offset variation since entering \texttt{GNSS}-Denied conditions, being unbiased (zero mean), bounded by atmospheric physics, and valid for all navigation purposes except landing.

\item The \texttt{VINS} \textbf{horizontal position} estimation exhibits drastic quantitative improvements over the baseline represented by the \texttt{INS} \cite{INSE}, which relies exclusively on an inertial filter without the use of onboard cameras and visual algorithms. Although from a qualitative point of view the estimation error is not bounded as the drift can not be fully eliminated, the horizontal position \texttt{VINSE} in the two evaluated scenarios is just a small fraction of the corresponding \texttt{INSE}. The reduced drift implies that the autonomous aircraft has a very high probability of at least reaching the vicinity of the predetermined recovery location, from where it can be landed by remote control, even if it requires flying for a long time in \texttt{GNSS}-Denied conditions while executing different types of maneuvers (tuns, climbs or descents, airspeed changes) in the presence of turbulent and varying weather.

The improvement can be attributed to the two main fusion steps: using the accurate body attitude inertial estimations to limit the drift in the visual estimations \cite{VNSE}, and simultaneously employing the ensuing more accurate horizontal position visual estimations to further reduce the inertial attitude errors (sections \ref{sec:VA-INS} and \ref{sec:VINS}). The approach hence results in a virtuous loop in which both \texttt{VINS} components (\texttt{IA-VNS} and \texttt{VA-INS}) simultaneously assist and control each other. 
\end{itemize}


\section{Conclusions} \label{sec:Conclusion}

The proposed visual inertial navigation filter, specifically designed for the challenges faced by autonomous fixed wing aircraft that encounter \texttt{GNSS}-Denied conditions, merges the observations provided by a suite of onboard sensors with those of the virtual vision sensor or \texttt{VVS}, resulting in major reductions in the horizontal position drift  that increase the possibilities of the platform successfully reaching the vicinity of a recovery location from where it can be landed by remote control. The \texttt{VVS} is the denomination of the outputs generated by a visual inertial odometry pipeline that relies on the images of the Earth surface generated by an onboard camera as well as on the navigation filter outputs. The filter is implemented in the manifold of rigid body rotations or special orthogonal group of \nm{\mathbb{R}^3}, known as \nm{\mathbb{SO}(3)}, to minimize the accumulation of errors in the absence of the absolute position observations provided by the \texttt{GNSS} receiver.

\appendix 


\section{GNSS-Denied Navigation and Visual Inertial Odometry}\label{sec:GNSS-Denied}

The number, variety, and applications of Unmanned Air Vehicles (\texttt{UAV}s) have grown exponentially in the last few years, and the trend is expected to continue in the future \cite{Hassanalian2017,Shakhatreh2019}. A comprehensive review of small \texttt{UAV} navigation systems and the problems they face is included in \cite{Bijjahalli2020}. The use of Global Navigation Satellite Systems (\texttt{GNSS}) constitutes one of the main enablers for autonomous inertial navigation\footnote{Inertial navigation is that which relies on the measurements provided by accelerometers and gyroscopes, also known as the Inertial Measurement Unit or \texttt{IMU}.} \cite{Farrell2008, Groves2008, Chatfield1997}, but it is also one of its main weaknesses, as in the absence of \texttt{GNSS} signals inertial systems are forced to rely on dead reckoning, which results in position drift, with the aircraft slowly but steadily deviating from its intended route \cite{Elbanhawi2017}. The availability of \texttt{GNSS} signals cannot be guaranteed; a throughout analysis of \texttt{GNSS} threats and reasons for signal degradation is presented in \cite{Sabatini2017}. 

At this time there are no comprehensive solutions to the operation of autonomous \texttt{UAV}s in \texttt{GNSS}-Denied scenarios, for which the permanent loss of the \texttt{GNSS} signals is equivalent to losing the airframe in an uncontrolled way. A summary of the challenges of \texttt{GNSS}-Denied navigation and the research efforts intended to improve its performance are provided by \cite{INSE, Tippitt2020, Gyagenda2022}. Two promising techniques for completely eliminating the position drift are the use of \emph{signals of opportunity}\footnote{Existing signals originally intended for other purposes, such as those of television and cellular networks, can be employed to triangulate the aircraft position.} \cite{Kapoor2017, Coluccia2014, Goh2013}, and \emph{georegistration}\footnote{The position drift can be eliminated by matching landmarks or terrain features as viewed from the aircraft to preloaded data.} \cite{Couturier2020, Goforth2019, Ziaei2019, Wang2016}, also known as \emph{image registration}.

Visual Odometry (\texttt{VO}) consists on employing the ground images generated by one or more onboard cameras without the use of prerecorded image databases or any other sensors, incrementally estimating the vehicle pose (position plus attitude) based on the changes that its motion induces on the images \cite{Scaramuzza2011, Fraundorfer2012, Scaramuzza2012, Scaramuzza2017, Cadena2016}. It  requires sufficient illumination, dominance of static scene, enough texture, and scene overlap between consecutive images or frames. Modern standalone algorithms such as Semi Direct Visual Odometry (\texttt{SVO}) \cite{Forster2014, Forster2016}, Direct Sparse Odometry (\texttt{DSO}) \cite{Engel2018}, Large Scale Direct Simultaneous Localization and Mapping (\texttt{LSD}-\texttt{SLAM}) \cite{Engel2014}, and large scale feature based \texttt{SLAM} (\texttt{ORB}-\texttt{SLAM})\footnote{\texttt{ORB} stands for Oriented \texttt{FAST} and rotated \texttt{BRIEF}, a type of blob feature.} \cite{Mur2015, Mur2017, Mur2017bis} are robust and exhibit a limited drift. Although it has been employed for navigation of ground robots, road vehicles, and multirotors flying both indoors and outdoors, the incremental concatenation of relative poses results in a slow but unbounded pose drift that disqualifies \texttt{VO} for long term autonomous \texttt{UAV} \texttt{GNSS}-Denied navigation.

A promising research path to \texttt{GNSS}-Denied navigation appears to be the fusion of inertial and visual odometry algorithms into an integrated navigation system. Two different trends coexist in the literature, although an implementation of either group can sometimes employ a technique from the other:
\begin{itemize}
\item The first possibility, known as \emph{filtering}, consists on employing the visual estimations as additional observations with which to feed the inertial filter. In the case of attitude estimation exclusively (momentarily neglecting the platform position), the \texttt{GNSS} signals are helpful and enable the \texttt{INS} to obtain more accurate and less noisy estimations, but they are not indispensable, as driftless attitude estimations can be obtained based on the \texttt{IMU} without their use. Proven techniques for attitude estimation in all kinds of platforms that do not rely on \texttt{GNSS} signals include Gaussian filters \cite{Crassidis2003}, deterministic filters \cite{Grip2012}, complimentary filters \cite{Kottah2017}, and stochastic filters \cite{Hashim2020, Hashim2019}. In the case of the complete pose estimation (both attitude and position), it is indispensable to employ the velocity or incremental position observations obtained with \texttt{VO} methods in the absence of the absolute references provided by \texttt{GNSS} receivers. The literature includes cases based on Gaussian filters \cite{Batista2011}, and more recently nonlinear deterministic filters \cite{Hashim2019b}, stochastic filters \cite{Hashim2021}, Riccati observers \cite{Hua2018}, and invariant extended Kalman filters or \texttt{EKF}s \cite{Barrau2017}.

\item The alternative is to employ \texttt{VO} optimizations with the assistance of the inertial navigation estimations, reducing the pose estimation drift inherent to \texttt{VO}. This is known as Visual Inertial Odometry (\texttt{VIO}) \cite{Scaramuzza2019, Huang2019}, which can also be combined with image registration to fully eliminate the remaining pose drift.  Current \texttt{VIO} implementations are also primarily intended for ground robots, multirotors, and road vehicles. \texttt{VIO} has matured significantly in the last few years, with detailed reviews available in \cite{Scaramuzza2019, Huang2019, Stumberg2019, Feng2019, Alkaff2017}.
\end{itemize}

There exist several open source \texttt{VIO} packages, such as the Multi State Constraint Kalman Filter (\texttt{MSCKF}) \cite{Mourikis2007}, the Open Keyframe Visual Inertial \texttt{SLAM} (\texttt{OKVIS}) \cite{Leutenegger2013,Leutenegger2015}, the Robust Visual Inertial Odometry (\texttt{ROVIO}) \cite{Bloesch2015}, the monocular Visual Inertial Navigation System (\texttt{VINS-Mono}) \cite{Qin2017}, \texttt{SVO} combined with Multi Sensor Fusion (\texttt{MSF}) \cite{Forster2014, Forster2016, Lynen2013, Faessler2016}, and \texttt{SVO} combined with Incremental Smoothing and Mapping (\texttt{iSAM}) \cite{Forster2014, Forster2016, Forster2016c, Kaess2012}. All these open source pipelines are compared in \cite{Scaramuzza2019}, and their results when applied to the EuRoC \texttt{MAV} data sets \cite{Burri2016} are discussed in \cite{Delmerico2018}. There also exist various other published \texttt{VIO} pipelines with implementations that are not publicly available \cite{Mur2017c, Clark2017, Paul2017, Song2017, Solin2017, Houben2016, Eckenhoff2017}, and there are also others that remain fully proprietary.

The existing \texttt{VIO} schemes to fuse the visual and inertial measurements can be broadly grouped into two paradigms: \emph{loosely coupled} pipelines process the measurements separately, resulting in independent visual and inertial pose estimations, which are then fused to get the final estimate; on the other hand, \emph{tightly coupled} methods compute the final pose estimation directly from the tracked image features and the \texttt{IMU} outputs \cite{Scaramuzza2019, Huang2019}. Tightly coupled approaches usually result in higher accuracy, as they use all the information available and take advantage of the \texttt{IMU} integration to predict the feature locations in the next frame. Loosely coupled methods, although less complex and more computationally efficient, lose information by decoupling the visual and inertial constraints, and are incapable of correcting the drift present in the visual estimator.

A different classification involves the number of images involved in each estimation \cite{Scaramuzza2019, Huang2019, Strasdat2010}, which is directly related with the resulting accuracy and computing demands. \emph{Batch algorithms}, also known as \emph{smoothers}, estimate multiple states simultaneously by solving a large nonlinear optimization problem or bundle adjustment, resulting in the highest possible accuracy. Valid techniques to limit the required computing resources include the reliance on a subset of the available frames (known as \emph{keyframes}\footnote{Common criteria to identify keyframes involve thresholds on the time, position, or pose increments from the previous keyframe.}), the separation of tracking and mapping into different threads, and the development of incremental smoothing techniques based on factor graphs \cite{Kaess2012}. Although employing all available states (\emph{full smoothing}) is sometimes feasible for very short trajectories, most pipelines rely on \emph{sliding window} or \emph{fixed lag smoothing}, in which the optimization relies exclusively on the measurements associated to the last few keyframes, discarding both the old keyframes as well as all other frames that have not been cataloged as keyframes. On the other hand, \emph{filtering algorithms} restrict the estimation process to the latest state; they require less resources but suffer from permanently dropping all previous information and a much harder identification and removal of outliers, both of which lead to error accumulation or drift.


\section{Required Jacobians} \label{sec:jacob}

This appendix groups together the various Jacobians employed in this article. They are the following:
\begin{itemize}

\item The time derivative of the geodetic coordinates \nm{\TEgdt = \lrsb{\lambda \ \phi \ h}^T} (longitude, latitude, and altitude) depends on the \texttt{NED} velocity \nm{\vN} per (\ref{eq:eq_TEgdt}), where M and N represent the \texttt{WGS84} ellipsoid radii of curvature of meridian and prime vertical, respectively. The Jacobian with respect to \nm{\vN}, given by (\ref{eq:jac_TEgdt_vN}), is hence straightforward:
\begin{eqnarray}
\nm{\TEgdtdot} & = & \nm{\lrsb{\dfrac{\vNii}{\lrsb{N\lrp{\varphi} + h} \, \cos\varphi} \ \ \ \ \dfrac{\vNi}{M\lrp{\varphi} + h} \ \ \ \ - \vNiii}^T} \label{eq:eq_TEgdt} \\
\nm{\vec J_{\ds{\; \vN}}^{\ds{\; \vec {\dot{T}}^{\sss E,GDT}}}} & = & \nm{\begin{bmatrix} 0 & \nm{\dfrac{1}{\lrp{N + h} \, \cos\varphi}} & 0 \\ \nm{\dfrac{1}{M + h}} & 0 & 0 \\ 0 & 0 & \nm{- 1} \end{bmatrix}  \ \ \in \mathbb{R}^{3x3}} \label{eq:jac_TEgdt_vN} 
\end{eqnarray}

\item The motion angular velocity \nm{\wEN = \dot\lambda \ \iEiii - \dot\varphi \ \iNii} represents the rotation experienced by any object that moves without modifying its attitude with respect to the Earth surface. It is caused by the curvature of the Earth and its expression when viewed in \texttt{NED} is given by (\ref{eq:eq_wENN}). Its Jacobian with respect to \nm{\vN}, provided by (\ref{eq:jac_wENN_vN}), is also straightforward:
\begin{eqnarray}
\nm{\wENN} & = & \nm{\lrsb{\dfrac{\vNii}{N\lrp{\varphi} + h} \ \ \ \ \dfrac{- \; \vNi}{M\lrp{\varphi} + h} \ \ \ \ \dfrac{- \; \vNii \; \tan \varphi}{N\lrp{\varphi} + h} }^T} \label{eq:eq_wENN} \\
\nm{\vec J_{\ds{\; \vN}}^{\ds{\; \wENN}}} & = & \nm{\begin{bmatrix} 0 & \nm{\dfrac{1}{N + h}} & 0 \\ \nm{\dfrac{-1}{M + h}} & 0 & 0 \\ 0 & \nm{\dfrac{- \tan \varphi}{N + h}} & 0 \end{bmatrix} \ \ \in \mathbb{R}^{3x3}} \label{eq:jac_wENN_vN}
\end{eqnarray}

\item The Coriolis acceleration \nm{\acor = 2 \ \wIE \ \vec v} is the double cross product between the Earth angular velocity caused by its rotation around the \nm{\iEiii} axis at a constant rate \nm{\omegaE} and the aircraft velocity. Its expression when viewed in \texttt{NED} is provided by (\ref{eq:eq_acorN}), resulting in the (\ref{eq:jac_acorN_vN}) Jacobian with respect to \nm{\vN}:

\begin{eqnarray}
\nm{\acorN} & = & \nm{2 \; \wIENskew \; \vN = 2 \; \omegaE \lrsb{\vNii \; \sin \varphi \ \ \ \ \lrp{- \vNi \; \sin \varphi - \vNiii \; \cos \varphi} \ \ \ \ \vNii \; \cos \varphi}^T} \label{eq:eq_acorN} \\
\nm{\vec J_{\ds{\; \vN}}^{\ds{\; \acorN}}} & = & \nm{2 \; \omegaE \; \begin{bmatrix} 0 & \sin \varphi & 0 \\ - \sin \varphi & 0 & - \cos \varphi \\ 0 & \cos \varphi & 0 \end{bmatrix}  \ \ \in \mathbb{R}^{3x3}} \label{eq:jac_acorN_vN}
\end{eqnarray}

\item The Lie Jacobian \nm{\vec J_{\ds{\; \mathcal R}}^{\ds{\; \mathcal R \oplus \Delta \vec r}}} represents the derivative of the function \nm{f\lrp{\mathcal R, \Delta\vec r} = \mathcal R \oplus \Delta\vec r}, this is, the concatenation between the \nm{\mathbb{SO}\lrp{3}} attitude \nm{\mathcal R} and its local perturbation \nm{\Delta\vec r}, with respect to the \nm{\mathbb{SO}\lrp{3}} attitude \nm{\mathcal R}, when the increments are viewed in their respective local tangent spaces, this is, tangent respectively at \nm{\mathcal R} and \nm{f\lrp{\mathcal R} = \mathcal R \oplus \Delta\vec r}. Interested readers should refer to \cite{LIE, Sola2018} for the obtainment of (\ref{eq:jac_lie1}), where \nm{\vec R\lrp{\Delta\vec r}} represents the direct cosine matrix corresponding to a given rotation vector \nm{\Delta\vec r}, and the wide hat \nm{< \widehat{\cdot} >} refers to the skew-symmetric form of a vector:
\neweq{\vec J_{\ds{\; \mathcal R}}^{\ds{\; \mathcal R \oplus \Delta \vec r}} = \vec R^T\lrp{\Delta \vec r} = \vec R\lrp{- \Delta \vec r} = \vec I_3 - \frac{\Delta \widehat{\vec r}}{\| \Delta \vec r \|} \sin \| \Delta \vec r \| + \frac{{(\Delta \widehat{\vec r}})^2}{\| \Delta \vec r \|^2} \lrp{1 - \cos \| \Delta \vec r \|} \ \ \ \in \mathbb{R}^{3x3}} {eq:jac_lie1}

\item The Lie Jacobian \nm{\vec J_{\ds{\; \mathcal R}}^{\ds{\; \vec g_{\mathcal R*}(\vec v)}}} represents the derivative of the function \nm{f\lrp{\mathcal R, \vec v} = \vec g_{\mathcal R*}(\vec v)}, this is, the rotation of \nm{\vec v} according to the \nm{\mathbb{SO}\lrp{3}} attitude \nm{\mathcal R}, with respect to the \nm{\mathbb{SO}\lrp{3}} attitude \nm{\mathcal R}, when the \nm{\mathcal R} increment is viewed in its local tangent space and that of the resulting \nm{\mathbb{R}^3} vector is viewed in its Euclidean space. \nm{\vec J_{\ds{\; \vec v}}^{\ds{\; \vec g_{\mathcal R*}(\vec v)}}} is the derivative of the same function with respect to the \nm{\mathbb{R}^3} unrotated vector \nm{\vec v}, in which the increments of the unrotated vector are also viewed in the \nm{\mathbb{R}^3} Euclidean space. Refer to \cite{LIE, Sola2018} for the obtainment of (\ref{eq:jac_lie2}) and (\ref{eq:jac_lie3}), where \nm{\vec R\lrp{\mathcal R}} represents the direct cosine matrix corresponding to the rotation \nm{\mathcal R}:
\begin{eqnarray}
\nm{\vec J_{\ds{\; \mathcal R}}^{\ds{\; \vec g_{\mathcal R*}(\vec v)}}} & = & \nm{- \vec R\lrp{\mathcal R} \, \widehat{\vec v} \ \ \in \mathbb{R}^{3x3}} \label{eq:jac_lie2} \\
\nm{\vec J_{\ds{\; \vec v}}^{\ds{\; \vec g_{\mathcal R*}(\vec v)}}} & = & \nm{\vec R\lrp{\mathcal R} \ \ \in \mathbb{R}^{3x3}} \label{eq:jac_lie3}
\end{eqnarray}

\item The Lie Jacobians \nm{\vec J_{\ds{\; \mathcal R}}^{\ds{\; \vec g_{\mathcal R*}^{-1}(\vec v)}}} and \nm{\vec J_{\ds{\; \vec v}}^{\ds{\; \vec g_{\mathcal R*}^{-1}(\vec v)}}} are similar to the previous ones but refer to the inverse rotation action \nm{f\lrp{\mathcal R, \vec v} = \vec g_{\mathcal R*}^{-1}(\vec v)}. Expressions (\ref{eq:jac_lie4}) and (\ref{eq:jac_lie5}) are also obtained in \cite{LIE, Sola2018}, where the wedge symbol \nm{< \cdot^{\wedge} >} refers to the skew-symmetric form of a vector:
\begin{eqnarray}
\nm{\vec J_{\ds{\; \mathcal R}}^{\ds{\; \vec g_{\mathcal R*}^{-1}(\vec v)}}} & = & \nm{\Big(\vec R^T\lrp{\mathcal R} \, \vec v\Big)^\wedge  \ \ \in \mathbb{R}^{3x3}} \label{eq:jac_lie4} \\
\nm{\vec J_{\ds{\; \vec v}}^{\ds{\; \vec g_{\mathcal R*}^{-1}(\vec v)}}} & = & \nm{\vec R^T\lrp{\mathcal R} \ \ \in \mathbb{R}^{3x3}} \label{eq:jac_lie5}
\end{eqnarray}

\item The Lie Jacobian \nm{\vec J_{\ds{\; \mathcal R}}^{\ds{\; \vec {Ad}_{\mathcal R}^{-1}(\vec \omega)}}} represents the derivative of the function \nm{f\lrp{\mathcal R, \vec\omega} = \vec {Ad}_{\mathcal R}^{-1}(\vec \omega)} , this is, the inverse of the adjoint of \nm{\vec\omega} according to the \nm{\mathbb{SO}\lrp{3}} attitude \nm{\mathcal R}, with respect to the \nm{\mathbb{SO}\lrp{3}} attitude \nm{\mathcal R}, when the \nm{\mathcal R} increment is viewed in its local tangent space and that of the resulting \nm{\mathbb{R}^3} vector is viewed in its Euclidean space. \nm{\vec J_{\ds{\; \vec \omega}}^{\ds{\; \vec {Ad}_{\mathcal R}^{-1}(\vec \omega)}}} is the derivative of the same function with respect to the \nm{\mathbb{R}^3} original vector \nm{\vec\omega}. Refer to \cite{LIE, Sola2018} for the obtainment of (\ref{eq:jac_lie6}) and (\ref{eq:jac_lie7}):
\begin{eqnarray}
\nm{\vec J_{\ds{\; \mathcal R}}^{\ds{\; \vec {Ad}_{\mathcal R}^{-1}(\vec \omega)}}} & = & \nm{\lrp{\vec R^T\lrp{\mathcal R} \, \vec\omega}^\wedge \ \ \in \mathbb{R}^{3x3}} \label{eq:jac_lie6} \\
\nm{\vec J_{\ds{\; \vec \omega}}^{\ds{\; \vec {Ad}_{\mathcal R}^{-1}(\vec \omega)}}} & = & \nm{\vec R^T\lrp{\mathcal R} \ \ \in \mathbb{R}^{3x3}} \label{eq:jac_lie7}
\end{eqnarray}

\end{itemize}


\bibliographystyle{ieeetr}   
\bibliography{visual_inertial}

\begin{thebibliography}{10}

\bibitem{CODE}
E.~Gallo, ``{High Fidelity Flight Simulation for an Autonomous Low SWaP Fixed
  Wing UAV in GNSS-Denied Conditions}.''
  \url{https://github.com/edugallogithub/gnssdenied_flight_simulation}, 2022.
\newblock C++ open source code.

\bibitem{LIE}
E.~Gallo, ``{The SO(3) and SE(3) Lie Algebras of Rigid Body Rotations and
  Motions and their Application to Discrete Integration, Gradient Descent
  Optimization, and State Estimation},'' 2022.
\newblock \texttt{arXiv:2205.12572v1 [cs.RO]},
  \url{https://doi.org/10.48550/arXiv.2205.12572}.

\bibitem{Sola2017}
J.~Sola, ``{Quaternion Kinematics for the Error-State Kalman Filter},'' 2017.
\newblock \texttt{arXiv:1711.02508v1 [cs.RO]},
  \url{https://doi.org/10.48550/arXiv.1711.02508}.

\bibitem{Sola2018}
J.~Sola, J.~Deray, and D.~Atchuthan, ``{A Micro Lie Theory for State Estimation
  in Robotics},'' 2018.
\newblock \texttt{arXiv:1812.01537v9 [cs.RO]},
  \url{https://doi.org/10.48550/arXiv.1812.01537}.

\bibitem{INSE}
E.~Gallo and A.~Barrientos, ``{Reduction of GNSS-Denied Inertial Navigation
  Errors for Fixed Wing Autonomous Unmanned Air Vehicles},'' {\em Aerospace
  Science and Technology}, 2021.
\newblock \url{https://doi.org/10.1016/j.ast.2021.107237}.

\bibitem{VNSE}
E.~Gallo and A.~Barrientos, ``{GNSS-Denied Semi Direct Visual Navigation for
  Autonomous UAVs Aided by PI-Inspired Priors},'' {\em Aerospace}, 2023.
\newblock \url{https://doi.org/10.3390/aerospace10030220}.

\bibitem{Forster2014}
C.~Forster, M.~Pizzoli, and D.~Scaramuzza, ``{SVO: Fast Semi-Direct Monocular
  Visual Odometry},'' in {\em {IEEE International Conference on Robotics and
  Automation}}, 2014.
\newblock \url{https://doi.org/10.1109/ICRA.2014.6906584}.

\bibitem{Forster2016}
C.~Forster, Z.~Zhang, M.~Gassner, M.~Werlberger, and D.~Scaramuzza, ``{SVO:
  Semidirect Visual Odometry for Monocular and Multicamera Systems},'' {\em
  {IEEE Transactions on Robotics}}, 2016.
\newblock \url{https://doi.org/10.1109/TRO.2016.2623335}.

\bibitem{SENSORS}
E.~Gallo and A.~Barrientos, ``{Customizable Stochastic High Fidelity Model of
  the Sensors and Camera onboard a Fixed Wing Autonomous Aircraft},'' {\em
  Sensors}, 2022.
\newblock \url{https://doi.org/10.3390/s22155518}.

\bibitem{Simon2006}
D.~Simon, {\em {Optimal State Estimation}}.
\newblock John Wiley \& Sons, 2006.
\newblock ISBN 0-471-70858-5.

\bibitem{Blanco2020}
J.~L. Blanco, ``{A Tutorial on SE(3) Transformation Parameterizations and
  On-Manifold Optimization},'' 2020.
\newblock \texttt{arXiv:2103.15980v2 [cs.RO]},
  \url{https://doi.org/10.48550/arXiv.2103.15980}.

\bibitem{SIMULATION}
E.~Gallo, ``{Stochastic High Fidelity Simulation and Scenarios for Testing of
  Fixed Wing Autonomous GNSS-Denied Navigation Algorithms},'' 2021.
\newblock \texttt{arXiv:2102.00883v3 [cs.RO]},
  \url{https://doi.org/10.48550/arXiv.2102.00883}.

\bibitem{Burri2016}
M.~Burri, J.~Nikolic, P.~Gohl, T.~Schneider, J.~Rehder, S.~Omari, M.~W.
  Achtelik, and R.~Siegwart, ``{The EuRoC MAV Datasets},'' {\em IEEE
  International Journal of Robotics Research}, 2016.
\newblock \url{https://doi.org/10.1177/0278364915620033}.

\bibitem{Delmerico2018}
J.~Delmerico and D.~Scaramuzza, ``{A Benchmark Comparison of Monocular
  Visual-Inertial Odometry Algorithms for Flying Robots},'' {\em IEEE
  International Conference on Robotics and Automation}, 2018.
\newblock \url{https://doi.org/10.1109/ICRA.2018.8460664}.

\bibitem{INSA}
E.~Gallo, ``{Quasi Static Atmospheric Model for Aircraft Trajectory Prediction
  and Flight Simulation},'' 2021.
\newblock \texttt{arXiv:2101.10744v1 [eess.SY]},
  \url{https://doi.org/10.48550/arXiv.2101.10744}.

\bibitem{Hassanalian2017}
M.~Hassanalian and A.~Abdelkefi, ``{Classifications, Applications, and Design
  Challenges of Drones: A Review},'' {\em Progress in Aerospace Sciences},
  2017.
\newblock \url{https://doi.org/10.1016/j.paerosci.2017.04.003}.

\bibitem{Shakhatreh2019}
H.~Shakhatreh, A.~H. Sawalmeh, A.~Al-Fuqaha, Z.~Dou, E.~Almaita, I.~Khalil,
  N.~S. Othman, A.~Khreishah, and M.~Guizani, ``{Unmanned Aerial Vehicles
  (UAVs): A Survey on Civil Applications and Key Research Challenges},'' {\em
  IEEE Access}, 2019.
\newblock \url{https://doi.org/10.1109/ACCESS.2019.2909530}.

\bibitem{Bijjahalli2020}
S.~Bijjahalli, R.~Sabatini, and A.~Gardi, ``{Advances in Intelligent and
  Autonomous Navigation Systems for Small UAS},'' {\em Progress in Aerospace
  Sciences}, 2020.
\newblock \url{https://doi.org/10.1016/j.paerosci.2020.100617}.

\bibitem{Farrell2008}
J.~A. Farrell, {\em {Aided Navigation, GPS with High Rate Sensors}}.
\newblock McGraw-Hill, Electronic Engineering Series, 2008.

\bibitem{Groves2008}
P.~D. Groves, {\em {Principles of GNSS, Inertial, and Multisensor Integrated
  Navigation Systems}}.
\newblock Artech House, GNSS Technology and Application Series, 2008.

\bibitem{Chatfield1997}
A.~B. Chatfield, {\em {Fundamentals of High Accuracy Inertial Navigation}}.
\newblock American Institute of Aeronautics and Astronautics, Progress in
  Astronautics and Aeronautics, vol 174, 1997.

\bibitem{Elbanhawi2017}
M.~Elbanhawi, A.~Mohamed, R.~Clothier, J.~Palmer, M.~Simic, and S.~Watkins,
  ``{Enabling Technologies for Autonomous MAV Operations},'' {\em Progress in
  Aerospace Sciences}, 2017.
\newblock \url{https://doi.org/10.1016/j.paerosci.2017.03.002}.

\bibitem{Sabatini2017}
R.~Sabatini, T.~Moore, and S.~Ramasamy, ``{Global Navigation Satellite Systems
  Performance Analysis and Augmentation Strategies in Aviation},'' {\em
  Progress in Aerospace Sciences}, 2017.
\newblock \url{https://doi.org/10.1016/j.paerosci.2017.10.002}.

\bibitem{Tippitt2020}
C.~Tippitt, A.~Schultz, and W.~Procino, ``{Vehicle Navigation: Autonomy Through
  GPS-Enabled and GPS-Denied Environments},'' state of the art report
  DSIAC-2020-1328, Defense Systems Information Analysis Center, December 2020.

\bibitem{Gyagenda2022}
N.~Gyagenda, J.~V. Hatilima, H.~Roth, and V.~Zhmud, ``{A Review of GNSS
  Independent UAV Navigation Techniques},'' {\em Robotics and Autonomous
  Systems}, 2022.
\newblock \url{https://doi.org/10.1016/j.robot.2022.104069}.

\bibitem{Kapoor2017}
R.~Kapoor, S.~Ramasamy, A.~Gardi, and R.~Sabatini, ``{UAV Navigation using
  Signals of Opportunity in Urban Environments: A Review},'' {\em Energy
  Procedia}, 2017.
\newblock \url{https://doi.org/10.1016/j.egypro.2017.03.156}.

\bibitem{Coluccia2014}
A.~Coluccia, F.~Ricciato, and G.~Ricci, ``{Positioning Based on Signals of
  Opportunity},'' {\em IEEE Communications Letters}, 2014.
\newblock \url{https://doi.org/10.1109/LCOMM.2013.123013.132297}.

\bibitem{Goh2013}
S.~T. Goh, O.~Abdelkhalik, and S.~A. Zekavat, ``{A Weighted Measurement Fusion
  Kalman Filter Implementation for UAV Navigation},'' {\em Aerospace Science
  and Technology}, 2013.
\newblock \url{https://doi.org/10.1016/j.ast.2012.11.012}.

\bibitem{Couturier2020}
A.~Couturier and M.~A. Akhloufi, ``{A Review on Absolute Visual Localization
  for UAV},'' {\em Robotics and Autonomous Systems}, 2020.
\newblock \url{https://doi.org/10.1016/j.robot.2020.103666}.

\bibitem{Goforth2019}
H.~Goforth and S.~Lucey, ``{GPS-Denied UAV Localization using Pre Existing
  Satellite Imagery},'' in {\em {IEEE International Conference on Robotics and
  Automation}}, IEEE, 2019.
\newblock \url{https://doi.org/10.1109/ICRA.2019.8793558}.

\bibitem{Ziaei2019}
N.~Ziaei, ``{Geolocation of an Aircraft using Image Registration Coupling Modes
  for Autonomous Navigation},'' 2019.
\newblock \texttt{arXiv:1909.02875v1 [eess.IV]},
  \url{https://doi.org/10.48550/arXiv.1909.02875}.

\bibitem{Wang2016}
T.~Wang, ``{Augmented UAS Navigation in GPS Denied Terrain Environments using
  Synthetic Vision},'' {\em Iowa State University Graduate Theses and
  Dissertations, 15835}, 2018.
\newblock \url{https://doi.org/10.31274/etd-180810-5462}.

\bibitem{Scaramuzza2011}
D.~Scaramuzza and F.~Fraundorfer, ``{Visual Odometry Part 1: The First 30 Years
  and Fundamentals},'' {\em IEEE Robotics \& Automation Magazine}, 2011.
\newblock \url{https://doi.org/10.1109/MRA.2011.943233}.

\bibitem{Fraundorfer2012}
F.~Fraundorfer and D.~Scaramuzza, ``{Visual Odometry Part 2: Matching,
  Robustness, Optimization, and Applications},'' {\em IEEE Robotics \&
  Automation Magazine}, 2012.
\newblock \url{https://doi.org/10.1109/MRA.2012.2182810}.

\bibitem{Scaramuzza2012}
D.~Scaramuzza, ``{Tutorial on Visual Odometry}.'' Robotics \& Perception Group,
  University of Zurich, 2012.
\newblock DOI not available.

\bibitem{Scaramuzza2017}
D.~Scaramuzza, ``{Visual Odometry and SLAM: Past, Present, and the Robust
  Perception Age}.'' Robotics \& Perception Group, University of Zurich, 2017.
\newblock DOI not available.

\bibitem{Cadena2016}
C.~Cadena, L.~Carlone, H.~Carrillo, Y.~Latif, D.~Scaramuzza, J.~Neira, I.~Reid,
  and J.~J. Leonard, ``{Past, Present, and Future of Simultaneous Localization
  and Mapping: Towards the Robust Perception Age},'' {\em IEEE Transactions on
  Robotics}, 2016.
\newblock \url{https://doi.org/10.1109/TRO.2016.2624754}.

\bibitem{Engel2018}
J.~Engel, V.~Koltun, and D.~Cremers, ``{Direct Sparse Odometry},'' {\em IEEE
  Transactions on Pattern Analysis and Machine Intelligence}, 2018.
\newblock \url{https://doi.org/10.1109/TPAMI.2017.2658577}.

\bibitem{Engel2014}
J.~Engel, T.~Schops, and D.~Cremers, ``{LSD-SLAM: Large Scale Direct Monocular
  SLAM},'' {\em European Conference on Computer Vision}, 2014.
\newblock \url{https://doi.org/10.1007/978-3-319-10605-2_54}.

\bibitem{Mur2015}
R.~Mur-Artal, J.~M.~M. Montiel, and J.~D. Tardos, ``{ORB-SLAM: a Versatile and
  Accurate Monocular SLAM System},'' {\em IEEE Transactions on Robotics}, 2015.
\newblock \url{https://doi.org/10.1109/TRO.2015.2463671}.

\bibitem{Mur2017}
R.~Mur-Artal and J.~D. Tardos, ``{ORB-SLAM2: an Open-Source SLAM System for
  Monocular, Stereo, and RGB-D Cameras},'' {\em IEEE Transactions on Robotics},
  2017.
\newblock \url{https://doi.org/10.1109/TRO.2017.2705103}.

\bibitem{Mur2017bis}
R.~Mur-Artal, {\em {Real-Time Accurate Visual SLAM with Place Recognition}}.
\newblock PhD thesis, University of Zaragoza, 2017.
\newblock \url{https://doi.org/10.1109/TRO.2017.2705103}.

\bibitem{Crassidis2003}
J.~L. Crassidis and F.~L. Markley, ``{Unscented Filtering for Spacecraft
  Attitude Estimation},'' {\em Journal of Guidance, Control, and Dynamics},
  2003.
\newblock \url{https://doi.org/10.2514/2.5102}.

\bibitem{Grip2012}
H.~F. Grip, T.~I. Fossen, T.~A. Johansen, and A.~Saberi, ``{Attitude Estimation
  Using Biased Gyro and Vector Measurements with Time Varying Reference
  Vectors},'' {\em IEEE Transactions on Automatic Control}, 2012.
\newblock \url{https://doi.org/10.1109/TAC.2011.2173415}.

\bibitem{Kottah2017}
R.~Kottah, P.~Narkhede, V.~Kumar, V.~Karar, and S.~Poddar, ``{Multiple Model
  Adaptive Complementary Filter for Attitude Estimation},'' {\em Aerospace
  Science and Technology}, 2017.
\newblock \url{https://doi.org/10.1016/j.ast.2017.10.010}.

\bibitem{Hashim2020}
H.~A. Hashim, ``{Systematic Convergence of Nonlinear Stochastic Estimators on
  the Special Orthogonal Group SO(3)},'' {\em International Journal of Robust
  and Nonlinear Control}, 2020.
\newblock \url{https://doi.org/10.1002/rnc.4971}.

\bibitem{Hashim2019}
H.~A. Hashim, L.~J. Brown, and K.~McIsaac, ``{Nonlinear Stochastic Attitude
  Filters on the Special Orthogonal Group SO(3): Ito and Stratonovich},'' {\em
  IEEE Transactions on Systems, Man, and Cybernetics}, 2019.
\newblock \url{https://doi.org/10.1109/TSMC.2018.2870290}.

\bibitem{Batista2011}
P.~Batista, C.~Silvestre, and P.~Oliveira, ``{On the Observability of Linear
  Motion Quantities in Navigation Systems},'' {\em Systems \& Control Letters},
  2011.
\newblock \url{https://doi.org/10.1016/j.sysconle.2010.11.002}.

\bibitem{Hashim2019b}
H.~A. Hashim, L.~J. Brown, and K.~McIsaac, ``{Nonlinear Pose Filters on the
  Special Euclidean Group SE(3) with Guaranteed Transient and Steady State
  Performance},'' {\em IEEE Transactions on Systems, Man, and Cybernetics},
  2019.
\newblock \url{https://doi.org/10.1109/TSMC.2019.2920114}.

\bibitem{Hashim2021}
H.~A. Hashim, ``{GPS Denied Navigation: Attitude, Position, Linear Velocity,
  and Gravity Estimation with Nonlinear Stochastic Observer},'' {\em
  Proceedings of the 2021 American Control Conference}, 2021.
\newblock \url{https://doi.org/10.23919/ACC50511.2021.9482995}.

\bibitem{Hua2018}
M.-D. Hua and G.~Allibert, ``{Riccati Observer Design for Pose, Linear
  Velocity, and Gravity Direction Estimation Using Landmark Position and IMU
  Measurements},'' {\em IEEE Conference on Control Technology and
  Applications}, 2018.
\newblock \url{https://doi.org/10.1109/CCTA.2018.8511387}.

\bibitem{Barrau2017}
A.~Barrau and S.~Bonnabel, ``{The Invariant Extended Kalman Filter as a Stable
  Observer},'' {\em IEEE Transactions on Automatic Control}, 2017.
\newblock \url{https://doi.org/10.1109/TAC.2016.2594085}.

\bibitem{Scaramuzza2019}
D.~Scaramuzza and Z.~Zhang, ``{Visual-Inertial Odometry of Aerial Robots},''
  2019.
\newblock \texttt{arXiv:1906.03289v2 [cs.RO]},
  \url{https://doi.org/10.48550/arXiv.1906.03289}.

\bibitem{Huang2019}
G.~Huang, ``{Visual-Inertial Navigation: A Concise Review},'' 2019.
\newblock \texttt{arXiv:1906.02650v1 [cs.RO]},
  \url{https://doi.org/10.48550/arXiv.1906.02650}.

\bibitem{Stumberg2019}
L.~{von Stumberg}, V.~Usenko, and D.~Cremers, ``{Chapter 7 - A Review and
  Quantitative Evaluation of Direct Visual Inertial Odometry},'' in {\em
  Multimodal Scene Understanding} (M.~Y. Yang, B.~Rosenhahn, and V.~Murino,
  eds.), Academic Press, 2019.
\newblock \url{https://doi.org/10.1016/B978-0-12-817358-9.00013-5}.

\bibitem{Feng2019}
X.~Feng, Y.~Jiang, X.~Yang, M.~Du, and X.~Li, ``{Computer Vision Algorithms and
  Hardware Implementations: A Survey},'' {\em Integration, the VLSI Journal},
  2019.
\newblock \url{https://doi.org/10.1016/j.vlsi.2019.07.005}.

\bibitem{Alkaff2017}
A.~Al-Kaff, D.~Martin, F.~Garcia, A.~{de la Escalera}, and J.~Maria, ``{Survey
  of Computer Vision Algorithms and Applications for Unmanned Aerial
  Vehicles},'' {\em Expert Systems With Applications}, 2017.
\newblock \url{https://doi.org/10.1016/j.eswa.2017.09.033}.

\bibitem{Mourikis2007}
A.~I. Mourikis and S.~I. Roumeliotis, ``{A Multi-State Constraint Kalman Filter
  for Vision-aided Inertial Navigation},'' {\em Proceedings 2007 IEEE
  International Conference on Robotics and Automation}, 2007.
\newblock \url{https://doi.org/10.1109/ROBOT.2007.364024}.

\bibitem{Leutenegger2013}
S.~Leutenegger, P.~Furgale, V.~Rabaud, M.~Chli, K.~Konolige, and R.~Siegwart,
  ``{Keyframe Based Visual Inertial SLAM Using Nonlinear Optimization},'' {\em
  Robotics: Science and Systems}, 2013.
\newblock \url{https://doi.org/10.3929/ethz-b-000236658}.

\bibitem{Leutenegger2015}
S.~Leutenegger, S.~Lynen, M.~Bosse, R.~Siegwart, and P.~Furgale, ``{Keyframe
  Based Visual Inertial SLAM Using Nonlinear Optimization},'' {\em The
  International Journal of Robotics Research}, 2015.
\newblock \url{https://doi.org/10.1177/0278364914554813}.

\bibitem{Bloesch2015}
M.~Bloesch, S.~Omari, M.~Hutter, and R.~Siegwart, ``{Robust Visual Inertial
  Odometry Using a Direct EKF Based Approach},'' {\em International Conference
  of Intelligent Robot Systems}, 2015.
\newblock \url{https://doi.org/10.3929/ethz-a-010566547}.

\bibitem{Qin2017}
T.~Qin, P.~Li, and S.~Shen, ``{VINS-Mono: A Robust and Versatile Monocular
  Visual Inertial State Estimator},'' {\em IEEE Transactions on Robotics},
  2018.
\newblock \url{https://doi.org/10.1109/TRO.2018.2853729}.

\bibitem{Lynen2013}
S.~Lynen, M.~W. Achtelik, S.~Weiss, M.~Chli, and R.~Siegwart, ``{A Robust and
  Modular Multi Sensor Fusion Approach Applied to MAV Navigation},'' {\em
  International Conference of Intelligent Robot Systems}, 2013.
\newblock \url{https://doi.org/10.1109/IROS.2013.6696917}.

\bibitem{Faessler2016}
M.~Faessler, F.~Fontana, C.~Forster, E.~Mueggler, M.~Pizzoli, and
  D.~Scaramuzza, ``{Autonomous, Vision Based Flight and Live Dense 3D Mapping
  with a Quadrotor Micro Aerial Vehicle},'' {\em Journal of Field Robotics},
  2015.
\newblock \url{https://doi.org/10.1002/rob.21581}.

\bibitem{Forster2016c}
C.~Forster, L.~Carlone, F.~Dellaert, and D.~Scaramuzza, ``{On Manifold Pre
  Integration for Real Time Visual Inertial Odometry},'' {\em IEEE Transactions
  on Robotics}, 2017.
\newblock \url{https://doi.org/10.1109/TRO.2016.2597321}.

\bibitem{Kaess2012}
M.~Kaess, H.~Johannsson, R.~Roberts, V.~Ila, J.~Leonard, and F.~Dellaert,
  ``{iSAM2: Incremental Smoothing and Mapping Using the Bayes Tree},'' {\em The
  International Journal of Robotics Research}, 2012.
\newblock \url{https://doi.org/10.1177/0278364911430419}.

\bibitem{Mur2017c}
R.~Mur-Artal and J.~M.~M. Montiel, ``{Visual Inertial Monocular SLAM with Map
  Reuse},'' {\em IEEE Robotics and Automation Letters}, 2017.
\newblock \url{https://doi.org/10.1109/LRA.2017.2653359}.

\bibitem{Clark2017}
R.~Clark, S.~Wang, H.~Wen, A.~Markham, and N.~Trigoni, ``{VINet:
  Visual-Inertial Odometry as a Sequence-to-Sequence Learning Problem},'' {\em
  Proceedings of the AAAI Conference on Artificial Intelligence}, 2017.
\newblock \url{https://ojs.aaai.org/index.php/AAAI/article/view/11215}.

\bibitem{Paul2017}
M.~K. Paul, K.~Wu, J.~A. Hesch, E.~D. Nerurkar, and S.~I. Roumeliotis, ``{A
  Comparative Analysis of Tightly Coupled Monocular, Binocular, and Stereo
  VINS},'' {\em IEEE International Conference on Robotics and Automation},
  2017.
\newblock \url{https://doi.org/10.1109/ICRA.2017.7989022}.

\bibitem{Song2017}
Y.~Song, S.~Nuske, and S.~Scherer, ``{A Multi Sensor Fusion MAV State
  Estimation from Long Range Stereo, IMU, GPS, and Barometric Sensors},'' {\em
  Sensors}, 2017.
\newblock \url{https://doi.org/10.3390/s17010011}.

\bibitem{Solin2017}
A.~Solin, S.~Cortes, E.~Rahtu, and J.~Kannala, ``{PIVO: Probabilistic Inertial
  Visual Odometry for Occlusion Robust Navigation},'' {\em IEEE Winter
  Conference on Applications of Computer Vision}, 2018.
\newblock \url{https://doi.org/10.1109/WACV.2018.00073}.

\bibitem{Houben2016}
S.~Houben, J.~Quenzel, N.~Krombach, and S.~Behnke, ``{Efficient Multi Camera
  Visual Inertial SLAM for Micro Aerial Vehicles},'' {\em IEEE/RSJ
  International Conference on Intelligent Robots and Systems}, 2016.
\newblock \url{https://doi.org/10.1109/IROS.2016.7759261}.

\bibitem{Eckenhoff2017}
K.~Eckenhoff, P.~Geneva, and G.~Huang, ``{Direct Visual Inertial Navigation
  with Analytical Preintegration},'' {\em IEEE International Conference on
  Robotics and Automation}, 2017.
\newblock \url{https://doi.org/10.1109/ICRA.2017.7989171}.

\bibitem{Strasdat2010}
H.~Strasdat, J.~M.~M. Montiel, and A.~J. Davison, ``{Real Time Monocular SLAM:
  Why Filter?},'' {\em IEEE International Conference on Robotics and
  Automation}, 2010.
\newblock \url{https://doi.org/10.1109/ROBOT.2010.5509636}.

\end{thebibliography}

\end{document}